\documentclass[10pt,twocolumn,letterpaper]{article}

\usepackage{cvpr}              % To produce the CAMERA-READY version
\definecolor{cvprblue}{rgb}{0.21,0.49,0.74}
\usepackage[pagebackref,breaklinks,colorlinks,allcolors=cvprblue]{hyperref}
\usepackage{mathtools}
\usepackage{multirow}
\usepackage{tcolorbox}
\tcbuselibrary{skins,breakable}
\usepackage{caption}
\usepackage{array}
\def\paperID{16957} % *** Enter the Paper ID here
\def\confName{CVPR}
\def\confYear{2026}

\def\inl{{\langle}}
\def\inr{{\rangle}}
\def\ienc{{\varphi}}
\def\tenc{{\psi}}
\def\beqa{\begin{eqnarray}}
\def\eeqa{\end{eqnarray}}
\def\beqann{\begin{eqnarray*}}
\def\eeqann{\end{eqnarray*}}
\def\loss{\ell}

\newcommand{\itt}{I$\rightarrow$T}
\newcommand{\tti}{T$\rightarrow$I}

\newcommand{\dataset}{{\textsc{Open-PMC}}\xspace}
\newcommand{\datasetsub}{{\textsc{Open-PMC-18M}}\xspace}
\newcommand{\datasetcomp}{{\textsc{PMC-6M}}\xspace}
\newcommand{\datasetsmall}{{\textsc{Open-PMC}}\xspace}
\newcommand{\biomedclip}{{\textsc{PMC-15M}}\xspace}

\newcommand{\biomedica}{\mbox{{\textsc{Biomedica}}}\xspace}

\newcommand{\negin}[1]{{\color{orange}[NB: #1]}}

\title{Open-PMC-18M: A High-Fidelity Large Scale Medical Dataset for Multimodal Representation Learning}
\author{
\begin{tabular}{c}
Negin Baghbanzadeh$^{1,2,*}$ \\
\end{tabular}
\hspace{1.5em}
\begin{tabular}{c}
Mohammed Saidul Islam$^{1,2,*}$ \\
\end{tabular}
\hspace{1.5em}
\begin{tabular}{c}
Sajad Ashkezari$^{1,3,*}$ \\
\end{tabular}
\\[0.7em]
\begin{tabular}{c}
Elham Dolatabadi$^{1,2}$ \\
\end{tabular}
\hspace{1.5em}
\begin{tabular}{c}
Arash Afkanpour$^{1}$ \\
\texttt{arash.afkanpour@vectorinstitute.ai}  % email only under his name
\end{tabular}
\\[1em]
\\
$^{1}$Vector Institute \quad
$^{2}$York University \quad
$^{3}$University of Waterloo \quad
$^{*}$Equal contribution
}

\begin{document}
% \twocolumn[{
% \begin{center}
% \includegraphics[width=\textwidth]{CVPR/figure/intro.pdf}
% \captionof{figure}{
% \textit{(a)} Compound figure from \datasetcomp dataset (\textbf{left}), and corresponding full caption and in-text reference related to the figure (\textbf{right}); 
% \textit{(b)} Extracted sub-figure image from \datasetsub (\textbf{top}), and corresponding extracted subcaption and summary of in-text reference related to the sub-figure image (\textbf{bottom}); 
% \textbf{(c)} Performance observations of training the proposed model in different settings.
% }
% \label{fig:intro}
% \end{center}
% }]
% \maketitle
\definecolor{highlight}{RGB}{255, 245, 170}
\definecolor{our_model}{RGB}{153, 102, 255}
\definecolor{other}{RGB}{135, 179, 255}
\twocolumn[{
  \renewcommand\twocolumn[1][]{#1}
  \maketitle
  \begin{center}
    \centering
    \captionsetup{type=figure}
    \includegraphics[width=\textwidth]{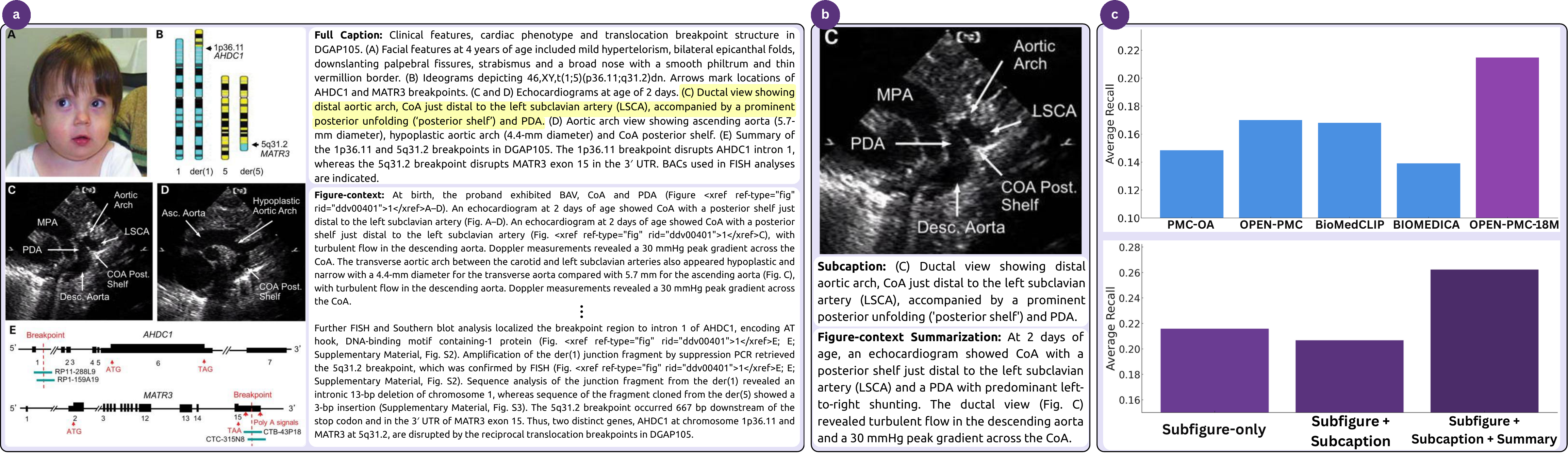}
    \captionof{figure}{\textit{(a)} Compound figure, and corresponding full caption (subcaption for subfigure `C' is \colorbox{highlight!80}{highlighted}) and in-text reference related to the figure from \biomedica dataset; \textit{(b)} Extracted subfigure image from \datasetsub, and corresponding extracted subcaption and summary of in-text reference; \textit{(c)} Average retrieval performance: \colorbox{our_model!80}{our model} vs. \colorbox{other!80}{other models} (\textbf{top}), and Retrieval results on MIMIC-CXR of model versions trained in different settings: \textit{subfigure only} vs. \textit{subfigure + subcaption} vs. \textit{subfigure + subcaption + summary} (\textbf{bottom}). The compound image in \textit{(a)} is originally from \citep{rivera2015matr}.}
    \label{fig:intro}
  \end{center}
}]
\begin{abstract}
In biomedical vision-language modeling, datasets are typically mined from scientific literature, pairing compound figures with captions that are short, context-dependent, and oftern partially informative. Prior work on subfigure extraction has been limited in both dataset size and generalizability. In addition, no existing effort has incorporated rich medical context in image-text pairs. We revisit data curation as a foundational component of effective biomedical representation learning. Our data curation process integrates transformer-based subfigure detection, subcaption extraction, and contextual text enrichment derived from inline references. Our subfigure extraction model, trained on a corpus of 500,000 compound figures, achieves state-of-the-art performance on real and synthetic benchmarks. Using this process, we curate and release \textbf{\datasetsub}, a large-scale high-fidelity biomedical dataset comprising \textit{18 million} image-text pairs, spanning radiology, microscopy, and visible light photography. We train vision-language models on our dataset and perform extensive evaluation on 6 retrieval and 19 zero-shot classification tasks across three major modalities. The models trained on our dataset set a new state-of-the-art results in medical representation learning. We release our dataset, models, and code to support reproducible benchmarks and further study into biomedical vision-language modeling and representation learning.
\end{abstract}

\section{Introduction}
\label{sec:intro}
The rapid progress of general-domain vision-language models (VLMs)~\citep{radford2021learning,jia2021scaling,girdhar2023imagebind} has inspired growing efforts to build large-scale multimodal datasets for scientific and biomedical representation learning~\citep{zhang2023biomedclip,lin2023pmc,pelka2018radiology,lozano2025biomedica,openpmc}.  
These models learn unified image-text representations that enable cross-modal retrieval and zero-shot classification, and can be further integrated with generative decoders to support downstream medical reasoning and report synthesis tasks \citep{li2023llava,saab2024capabilities}. 

A common approach to learn representations is via a contrastive pretraining objective, which aligns visual and textual embeddings in a shared semantic space \citep{oord2018representation,radford2021learning}. While broadly applicable, its effectiveness depends on the fidelity of the paired supervision~\citep{xu2024demystifying}. In natural web data, such as those used for CLIP~\citep{radford2021learning} and subsequent models~\citep{yao2021filip,mu2022slip}, the paired text (captions, metadata, or surrounding paragraphs) is typically descriptive and diverse, providing rich linguistic grounding for visual concepts~\citep{xu2024demystifying, gadre2023datacomp}. As later revealed by MetaCLIP~\citep{xu2024demystifying}, much of CLIP's success stems largely from meticulous data curation. This ensures that each image-text pair contributes semantically informative signals to the shared space.

In the biomedical domain, however, the paired supervision is constructed differently. Most medical VLMs are trained on datasets mined from scientific literature, such as PubMed Central (PMC)\footnote{https://pmc.ncbi.nlm.nih.gov/}(e.g, \biomedclip \citep{zhang2023biomedclip} and \biomedica \citep{lozano2025biomedica}), in which figures and captions are extracted to form image-text pairs (see Figure \ref{fig:intro}(a)). This introduces two fundamental challenges. First, biomedical figures are often \textit{compound} or multi-panel, combining heterogeneous content, including radiology scans, microscopy images, plots, and annotations. Second, the corresponding captions are frequently symbolic, abbreviated, or context-dependent, relying heavily on in-text references rather than self-contained descriptions. 
Pairing a compound image with a partially informative caption produces a data point that, when used for representation learning, leads to suboptimal representations. Yet, most existing datasets are formed in this manner.

These mismatches produce coarse image-text alignments during pretraining, introducing systemic errors into representation learning and encouraging models to fit to superficial or repetitive textual anchors rather than learning precise biomedical semantics. Over time, this weakens cross-modal correspondences and limits transfer to downstream tasks that demand fine-grained understanding. Thus, biomedical data amplify the lessons of MetaCLIP: high-fidelity VLMs require deliberate reconstruction of meaningful image–text pairs, not raw extraction of figures and captions.

%We hypothesize that this compoundness of biomedical figures and the abstract, underspecified nature of their accompanying text lead to coarse image-text alignment during pretraining. Such misalignment can introduce systematic noise into the contrastive objective, encouraging models to cluster around superficial or repetitive textual anchors rather than learning precise, semantically grounded biomedical representations. Over time, this weakens the learned cross-modal correspondences, ultimately limiting the transferability and generalizability of medical VLMs to downstream tasks that demand fine-grained biomedical understanding. %We hypothesize that such coarse image-text alignment could introduce noise into pretraining, ultimately impacting the transferability and generalizability of the learned representations.

One might argue that at sufficient data scale and model complexity, such error will vanish~\citep{sutton2019bitter}. However, scientific corpora such as PubMed Central, remain orders of magnitude smaller than open-web sources. Moreover, even in the natural domain, gains in zero-shot performance have been traced to curation and distributional balancing rather than sheer scale or architectural novelty \citep{xu2024demystifying,schuhmann2022laion,schuhmann2021laion}. This suggests that representation quality in medical VLMs is constrained more by data structure and alignment than by dataset size or model capacity.

% In this work we focus on careful extraction of subfigures and augmenting captions with inline text summaries.
To our knowledge, no prior work has jointly incorporated both subfigure extraction and contextual text summarization into a unified biomedical vision-language dataset, nor evaluated their impact on representation learning. A few prior works tackled subfigure extraction but at limited accuracy and scale \citep{pelka2018radiology,lin2023pmc,openpmc}; contextual enrichment has been largely unaddressed.

This raises an important gap in the field: \textit{how does subfigure extraction and contextual text summarization affect the quality of learned representations in the medical domain, particularly given the known sensitivity of contrastive objectives to dataset alignment and scale during pretraining?}

We introduce \datasetsub, the first large-scale and carefully curated dataset comprising 18 million subfigure-text pairs, where each text entry consists of an associated \emph{subcaption}, extracted from the original caption, and a summary of inline context that explicitly references the subfigure (Figure~\ref{fig:intro}(b)). Starting from the \biomedica corpus \citep{lozano2025biomedica}, which is extracted from PubMed Central, we apply metadata-level filtering using label annotations and a ResNet-based classifier to remove non-medical or schematic content. For subfigure extraction, we train a high-performance detector on a dataset of 500,000 programmatically generated compound figures. For subcaption extraction we utilized Qwen2.5-VL-32B-Instruct VLM \citep{Qwen2.5-VL} and text summarization we used Qwen2.5-14B-Instruct \citep{qwen2025qwen25technicalreport}.

We train vision and text encoders using contrastive loss on \datasetsub and evaluate the encoders on an extensive suite of downstream tasks, including cross-modal retrieval and zero-shot classification tasks across three major medical modalities: radiology, microscopy, and visible light photography (VLP). We release our dataset,
% \footnote{https://huggingface.co/datasets/vector-institute/open-pmc-18m}, 
% \footnote{https://anonymous.4open.science/r/open-pmc-18m-CE25/}
models, and code\footnote{https://github.com/vectorInstitute/pmc-data-extraction} to support reproducible benchmarks and further study into biomedical VLM and representation learning. Our contributions are as follows:
\begin{itemize}
    \item We curate and release \datasetsub, a large-scale high-fidelity biomedical image-text dataset with 18 million image-text pairs, each consists of a subfigure paired with the corresponding subcaption and image-context summary.
    \item We propose a scalable and accurate subfigure extraction pipeline based on transformer-based object detection, trained on 500,000 compound figures, achieving state-of-the-art performance on ImageCLEF 2016 \citep{KGD2014,GSB2016} and synthetic evaluation sets.
    \item We perform a comprehensive evaluation of VLMs trained on \datasetsub, demonstrating improved performance in retrieval (Figure~\ref{fig:intro} (c) (top)), classification, and robustness across multiple medical benchmarks in radiology, microscopy, and visible light photography.
\end{itemize}

\section{\datasetsub Composition and Curation Process}
% \section{\datasetsub dataset}
\label{sec:data}
Our dataset curation pipeline (Figure \ref{fig:layout} (a)) contain three key stages: \textit{(i)} Initial Figure Collection and Filtering, \textit{(ii)} Vision-Based Subfigure Extraction, and \textit{(iii)} Textual Enrichment via subcaption extraction and summary generation. Below we describe each stage in detail.

\subsection{Initial Figure Collection and Filtering}
\label{6m}
To curate our dataset we start with the \biomedica dataset \citep{lozano2025biomedica}, which has been extracted from articles in the PubMed Central Open Access Subset. \biomedica contains approximately 24 million image-caption pairs along with in-text figure references, often spanning multiple paragraphs. \biomedica obtains image modalities by extracting visual features of each image using a DINO-v2 model \citep{caron2021emerging}, followed by clustering of images with PCA and K-means. Clusters are then annotated by experts which provide 12 global modality labels (e.g., clinical image, microscopy, immunoassays, chemical structures) and 170 local labels (e.g., magnetic resonance imaging, X-ray radiography, computed tomography, etc.)

To improve dataset quality, we apply a filtering step using the provided labels and retain only those pairs primarily categorized as clinical imaging, microscopy, immunoassays, or chemical structure. This yields a dataset of 6 million pairs, which we refer to as \datasetcomp in this paper. %\saidul{In addition, we collected metadata fields, including both global and local image modalities, panel type (single or multiple-panel), as well as in-text references that often span multiple paragraphs. These metadata were annotated by a group of seven experts with backgrounds in genetics, pathology, surgery, developmental biology, and biomedical informatics, as described in \citep{lozano2025biomedica}}

\subsection{Vision-Based Subfigure Extraction} \label{detr}
To extract subfigures from biomedical compound figures with high accuracy at scale, we trained a transformer-based object detection model based on the \emph{Dynamic Anchor Box DEtection TRansformer} (DAB-DETR) architecture \citep{liu2022dabdetr}. Prior work of \citet{lin2023pmc} trained a DETR model for the same purpose on MedICaT \citep{subramanian2020medicat} with only 2,069 manually annotated compound figures. In contrast, we trained our model on a large-scale synthetic dataset of 500,000 compound figures, which we explain below. This dataset is the first of its kind in the biomedical domain. We use DAB-DETR as it improves upon the original DETR model by learning dynamic anchors as queries, resulting in improved localization and faster convergence \citep{liu2022dabdetr}.

%To enable automated extraction of individual subfigures from compound figures, we developed an object detection model based on the DEtection TRansformer (DETR) architecture. The objective was to locate and segment subfigures within larger, composite images typically found in scientific literature. Our approach reframes subfigure detection as a direct set prediction problem, where the model simultaneously predicts the number, location, and bounding boxes of subfigures without relying on hand-crafted proposals or post-processing steps.
\paragraph{Synthetic Data Formation.} To train a subfigure extraction model, we generate a synthetic dataset by reversing the subfigure extraction process: rather than decomposing existing compound figures, we programmatically construct new ones by composing multiple single-panel biomedical images into compound layouts. The key advantage of this approach is the availability of ground-truth bounding boxes for each subfigure. To create diverse layouts, our generation pipeline samples a layout template that specifies the spatial arrangement of subfigures from a large number of configurations. Each layout is defined by a set of configurable parameters, including:
% (Figure \ref{fig:layout}):
\begin{itemize}
\setlength\itemsep{0.1em}
\item \textbf{Grid Size}: Specifies a standard $m \times n$ grid or a custom arrangement for panel placement.
\item \textbf{Margins}: Random horizontal and vertical spacing between panels to simulate variability in published figure layouts.
\item \textbf{Labeling Scheme}: Determines how panels are annotated (e.g., using numerical, alphabetical, or compound labels like ``1a'' or ``a-1''), and whether labels appear inside or outside panel boundaries.
\item \textbf{Aspect Ratio}: Specifies a fixed width-to-height ratio applied uniformly to all subfigures.
\end{itemize}

\begin{figure*}[tb]
\centering
\includegraphics[width=0.99\textwidth]{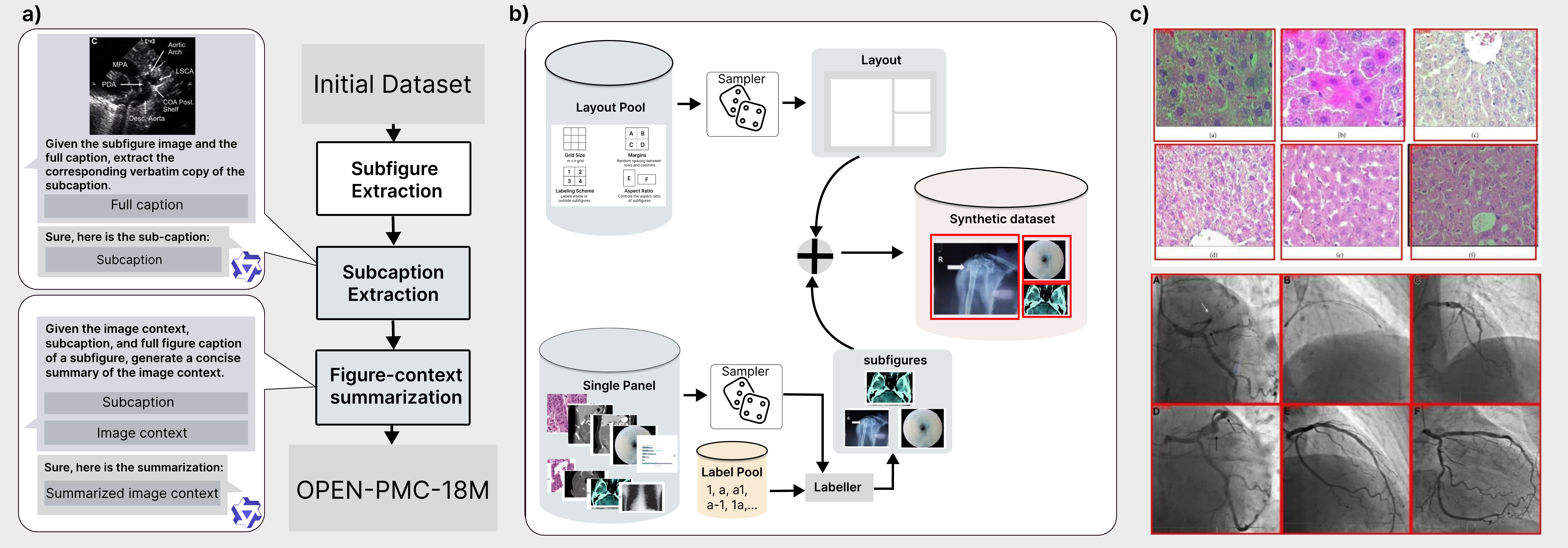}
\caption{\textit{(a)} Overview of the \datasetsub construction process comprising of the following key stages: \textit{(b)} Subfigure Extraction pipeline for creating synthetic compound figures that are used to train the DAB-DETR model. \textit{(c)} Example subfigures extracted using the DAB-DETR model from ImageCLEF dataset.}
% \caption{\elham{Needs to be revised; also a,b,c,d, is confusing}  \textbf{Left} Overview of our pipeline for creating synthetic compound figures used to train the DAB-DETR model. A 
% \textit{Sampler} selects single-panel images and layout specifications from their respective pools. A \textit{Labeler} assigns subfigure labels from a predefined label pool (e.g., 1, a, a1, a-1), placing them according to the chosen scheme. \textbf{Right} Distribution of medical image modalities, number of subfigures per compound figure, and caption length statistics within \datasetsub. The average caption contains 165.82 tokens, with a max of 7352 and almost 19.48\% of captions had more than 256 tokens. }
\label{fig:layout}
\end{figure*}

% \biomedica dataset also provides metadata describing the panel type of each images, such as whether it is a single-panel or multi-panel image. 
Subfigures are sampled from single-panel biomedical images spanning diverse modalities (also sourced from the metadata) such as radiology, microscopy, and VLP, which we will describe below. Composite figures may contain panels from the same modality or a heterogeneous mix, providing semantic diversity and mimicking real-world figure complexity. Figure \ref{fig:layout}(b)) illustrates the full synthetic data pipeline. 
% \arash{The referenced figure does not show a pipeline. Please fix.}\saidul{fixed}

\paragraph{Image Decomposition Model Training and Evaluation.}
We train a DAB-DETR model on the 500,000 compound figures and validate its performance on a similarly created holdout set of 20,000 images. Source subfigures are drawn from well-known benchmark datasets such as ROCO \citep{pelka2018radiology}, SICAP \citep{ESTEBAN2019303}, HAM10000 \citep{tschandl2018ham10000}, PathMNIST and RetinaMNIST from MedMNIST \citep{medmnistv1, medmnistv2}, PAD-UFES-20 \citep{pacheco2020pad}, and PlotQA \citep{Methani_2020_WACV} as listed in Table~\ref{tab:single_panel_datasets}. To ensure balanced representation, each modality-specific dataset contributes approximately 16.7\% of the total examples, with the remaining 16.7\% comprising mixed-modality compound figures. This configuration promotes both visual diversity and generalization across biomedical imaging types. Training is performed over $40$ epochs using a batch size of $64$ and an initial learning rate of $10^{-5}$. 

We evaluate performance on both our synthetic validation set and the ImageCLEF 2016 compound figure separation benchmark \citep{KGD2014, GSB2016}. For comparison, we benchmark against a DETR model trained on MedICaT~\citep{subramanian2020medicat} and the Qwen2.5-VL-32B-Instruct \citep{Qwen2.5-VL} multimodal model, which we adapt for zero-shot subfigure detection by prompting bounding-box prediction on compound figures. As shown in Table~\ref{tab:subfigure_evaluation}, our DAB-DETR substantially outperforms both baselines. 
Figure~\ref{fig:layout}(c) presents qualitative examples from the ImageCLEF 2016 dataset, and additional results from a subset of \datasetcomp\ are shown in Figure~\ref{fig:example} in Appendix \S\ref{sec:add_sub_ext}. Together, these examples demonstrate our model’s ability to accurately detect distinct subfigures across heterogeneous layouts and content types, in contrast to the inconsistent or missing predictions observed with Qwen2.5-VL-32B-Instruct (Figure~\ref{fig:qwenexample} in Appendix).

\begin{table}[h]
    \centering
    \caption{Single panel datasets used to construct the synthetic compound figure corpus, categorized by modality and split.
The training and validation sets are composed of single modality images that serve as source panels for synthetic figure generation.
$^{*}$ indicates the held-out test portion of ROCO, and $^{+}$ denotes the official validation split.}
    \setlength{\tabcolsep}{2pt} 
    \resizebox{\columnwidth}{!}{%
    \begin{tabular}{lcc}
        \hline
        \textbf{Modality} & \textbf{Train (Examples)}& \textbf{Validation (Examples)} \\
        \hline
        \textbf{Radiology} & ROCO (65422) & ROCO$^{*}$ (8176) \\
        \textbf{Histopathology} & SICAP (18783) & PathMNIST (10004) \\
        \textbf{Dermatology} & HAM10000 (10015) & PAD-UFES-20 (2298) \\
        \textbf{Retina} & RetinaMNIST (1080) & RetinaMNIST$^{+}$ (120) \\
        \textbf{Plots} & PlotQA (60000) & PlotQA$^{+}$ (10000) \\
        \hline
    \end{tabular}%
    }
    \label{tab:single_panel_datasets}
\end{table}

\begin{table}[h]
\centering
\setlength{\tabcolsep}{5pt}  % Adjust column spacing (default is 6pt)
\caption{Comparison of subfigure detection performance on two datasets using mAP and F1 metics. $^{*}$ Previous SOTA (MedICaT-trained DETR), $^{+}$ Qwen2.5-VL-32B-Instruct, and $^{o}$ our DAB-DETR model.}
\label{tab:subfigure_evaluation}
\begin{tabular}{lcc|cc}
\toprule
\textbf{Model} & \multicolumn{2}{c|}{\textbf{Synthetic Validation}} & \multicolumn{2}{c}{\textbf{ImageCLEF 2016}} \\
\cmidrule(lr){2-3} \cmidrule(lr){4-5}
 & \textbf{mAP (\%)} & \textbf{F1 (\%)} & \textbf{mAP (\%)} & \textbf{F1 (\%)} \\
\midrule
SOTA$^{*}$ & 33.22 & 73.18 & 28.20 & 64.85 \\
Qwen$^{+}$ & 40.31 & 79.01 & 30.12 & 68.12 \\
Ours$^{o}$    & \textbf{98.58} & \textbf{99.96} & \textbf{36.88} & \textbf{73.55} \\
\bottomrule
\end{tabular}
\label{tab: subfigure_evaluation}
\end{table}

\subsection{Text Enrichment}
To enrich the dataset with contextual information relevant to each subfigure, we employed two key methods: (a) We extracted the corresponding subcaption from the full caption to improve image-text alignment, and (b) We identified in-text reference paragraphs from the metadata and summarized them to further enrich the semantic content related to each subfigure.

\paragraph{Subcaption Extraction}
The majority of captions in the dataset consist of multiple subcaptions, generally distinguished by delimiters such as \textit{(a)}, \textit{(b)}, or \textit{(A–F)}. Given that the total number of subfigures is approximately 18 million, manual extraction of all subcaptions is not feasible. To address this, we adopted an automated extraction method using a state-of-the-art open-source VLM, i.e., Qwen2.5-VL-32B-Instruct \citep{Qwen2.5-VL}. 
% For each sample, we designed a prompt $P$ comprising the full caption $f$, a subfigure image $s$, and additional  instructions $I$ (see Figure~\ref{fig:subcap}) for subcaption extraction. 
For each sample, we designed a prompt comprising the full caption, a subfigure image, and additional instructions (Figure~\ref{fig:subcap} in the Appendix) for subcaption extraction.
% \arash{Are these symbols (P, f, ...) used anywhere? If not, remove them.} \saidul{no, removing.} 

To ensure determinism and prevent hallucinations or inaccuracies, the model was explicitly instructed to extract a verbatim copy of the subcaption directly from the full caption, without adding any additional text or explanation. In cases where a subcaption could not be found, the model was instructed to extract the verbatim copy of the full caption instead. Overall, we found that 13.28\% of subcaptions in the dataset are exact copies of their full captions. To evaluate the quality of our automated extraction method, we manually reviewed a random sample of 1,000 instances and observed that in 94\% of cases the subcaptions were correctly extracted. The erroneous cases were primarily due to missing subcaptions, indicating that the model performs reliably for large-scale automated subcaption extraction.
% \arash{Do we have the number of such cases? If yes, include it here.} To evaluate the quality of the automated extraction method, we utilized LLM-as-a-judge method using Qwen2.5-VL-32B-Instruct \arash{Does this mean the same Qwen model was used for both extraction and as a judge? Is it possible to run this evaluation with a different judge model?} \saidul{yes. Qwen is the best open-source VLM we have right now. Should we use other models for judging?} and evaluated a representative subset of 1,000 samples. We observed approximately 6\% error, primarily due to missing subcaptions, indicating that the model performs reliably for large-scale automated subcaption extraction. 

Inference was conducted on 32$\times$A100 GPUs (4 nodes), requiring roughly 13,474 GPU-hours (17.5 wall-clock days), with an average per-instance inference time of $~$3 seconds. 
% Figure~\ref{fig:subcap-judge} in the Appendix illustrates the judge prompt used for evaluation. 

\paragraph{In-text Reference Summarization}
Since the compound figures of \datasetsub were collected from the \biomedica dataset \citep{lozano2025biomedica}, we used the dataset’s accompanying image metadata, such as \emph{image context} which are inline text references associated with each compound figure. However, these references frequently extend across multiple paragraphs. In many cases, important contextual details appear outside the sentences that directly correspond to a subfigure, making it difficult to determine which pieces are related to a particular subfigure. To mitigate these issues, we employed the Qwen2.5-14B-Instruct \citep{qwen2025qwen25technicalreport} model to generate concise, context-aware summaries that distill the most relevant information while retaining essential details for each subfigure. For this task, we used this model instead of Qwen2.5-VL-32B-Instruct because the summary generation task involves only textual input and the 14B model is a text-only model optimized for instruction following and long-context generation \citep{qwen2025qwen25technicalreport}, making it a more suitable and computationally efficient option.\footnote{The 14B model takes on average 0.64s per instance as opposed to 3s for Qwen2.5-VL-32B-Instruct.}
% \arash{We're using a different qwen model here. Can we justify this?}
% For each sample, we designed a prompt $P$ comprising the full caption $f$, a subcaption related to a subfigure image $s$, the image context $c$ and additional  instructions $I$.
For each sample, we designed a prompt comprising the full caption, a subcaption related to a subfigure image, the image context, and additional instructions.
We provide the full prompt in Figure~\ref{fig:summary} in the Appendix (\S\ref{sec:prompts}).

\subsection{Final Refinement}
Decomposing the compound images of \datasetcomp using our DAB-DETR model yields an initial dataset of approximately 32 million. To ensure the final dataset contains only biomedical images, we apply an additional filtration using global modality metadata fields to only keep subfigures whose original compound figure was labeled as either Clinical Image or Microscopy, which yields a dataset of 26 million pairs. Subsequently, we employ a ResNet-101 model \citep{lin2023pmc} to assess each image and infer its medical relevance. This filtering process further reduces the dataset to 18 million high-quality image-caption pairs.

\subsection{Dataset Statistics}
\label{data_stats}
\datasetsub contains 12,997,862 microscopy images (73\%), 3,244,121 radiology images (18\%), 1,421,804 visible light photography images (8\%), and 204,212 other clinical imaging samples (1\%). Full caption lengths are highly variable, with most captions between 50 and 200 tokens, an average length of 165.8 tokens, and a maximum length of 7352; about 19.48\% of captions exceed 256 tokens. Subcaption lengths are relatively short and consistent across the dataset. The average subcaption contains 30 tokens with majority (82.91\%) containing less than 50 tokens. An additional 16.59\% of subcaptions contain between 50-150 tokens, and only 0.50\% exceed 150 tokens. Figure-context summaries are longer and demonstrate greater variability compared to subcaptions. The average summary length is 54 tokens. Less than half of summaries (46.21\%) contain fewer than 50 tokens, whereas 53.68\% are within the 50–150 token range. Only 0.11\% of summaries exceed 150 tokens. We limit the summaries to be less or equal to 256 tokens by limiting the maximum token generation to 256 tokens during model inference. In terms of figure structure, most figures contain between one and three subfigures, though some include up to 20. Detailed visualizations are provided in Figure \ref{fig:supp_dataset_statistics} in the \S\ref{sec:data_stats_supp} of the Appendix.
% \elham{Detailed visualizations are provided in Figure \ref{fig:supp_dataset_statistics} in the \S\ref{sec:data_stats_supp} of the supplementary material.}\saidul{updated}

\section{Experiments}
\label{sec:Experiment}

\subsection {Encoder Pretraining}
We train separate image and text encoders by aligning their representations using a vanilla contrastive loss~\citep{oord2018representation}. See \S\ref{subsec:enc-pretrain} for more details.

\subsection{Evaluation Setup}\label{eval}
To systematically assess the impact of dataset scale and curation quality, we perform evaluations along both dimensions. We train all models under a unified architecture and training protocol to ensure controlled evaluation. For models without accessible training data, we instead use publicly released checkpoints obtained from HuggingFace. For the text encoder, we use PubMedBERT \citep{pubmedbert}, and for the vision encoder, we adopt a ViT-B/16 transformer \citep{dosovitskiy2020image} pretrained on ImageNet. Encoders are trained for 60 epochs with a batch size of 2,048, and the best performing checkpoints are selected according to cross-modal retrieval performance on a held-out validation subset of 50,000 randomly sampled pairs. Details of the validation procedure and selection criteria are provided in the Appendix (\S\ref{val}). Image–text pairs used for training consist of subfigures, the extracted subcaptions, and summarized inline text. Training was performed on 8$\times$A100 GPUs and completed in five days. We conducted our experiments using the open-source \emph{mmlearn} multimodal learning framework.\footnote{\url{https://github.com/VectorInstitute/mmlearn/tree/main}}

To establish a baseline, we train VLM encoders on the \datasetcomp dataset, where each image-text pair consists of a compound image in its original form (Section \ref{6m}). We also include publicly available checkpoints from other models trained on \biomedclip \citep{zhang2023biomedclip} and \biomedica \citep{lozano2025biomedica}. For \biomedica, we use the checkpoint referred to as \texttt{BMC-CLIP}$_\text{CF}$ in \citet{lozano2025biomedica}, which is trained on a filtered subset of the full dataset, achieving state-of-the-art performance at the time of publishing \biomedica \citep{lozano2025biomedica}.
% This subset retains content labeled under clinical and scientific imaging, immunoassays, illustrative diagrams, chemical structures, maps, tools and materials, and hand-drawn or screen-based visuals, while explicitly excluding tables and charts.
For \biomedclip, we use the checkpoint trained on 15 million image-caption pairs, referred to as \texttt{BioMedCLIP} in \citet{zhang2023biomedclip}. All external checkpoints were obtained from their official HuggingFace repositories and are evaluated using our standard downstream protocols.

To further ensure consistency, we independently reproduce the PMC-OA dataset \citep{lin2023pmc} and train encoders using the same architecture and hyperparameters as those used for \datasetsub and \datasetcomp. Throughout the paper, all encoder variants are referenced by the name of the dataset on which they are trained, to facilitate transparent comparison. All the details of pretraining and hyperparameters are listed in the \S\ref{sec:train_det_supp} in the Appendix.
% \arash{Mention which section}

\begin{table*}[h]
\centering
\caption{Retrieval performance (Recall@200) of all models trained on paired image-caption pairs in the medical domain. The last column, Average Recall (AR), aggregates the results across all tasks. Highest performance values are in bold, second-best are underlined. \datasetcomp refers to a baseline model trained on a filtered subset of the \biomedica dataset, using compound figures in their original form without subfigure decomposition. The \biomedica model retrieved from HuggingFace is trained on a filtered subset of the full dataset, as described in their original paper.}
{\small
\begin{tabular}{lccc|ccc|c}
\toprule
              & \multicolumn{3}{c|}{\textbf{Image-to-Text}} & \multicolumn{3}{c|}{\textbf{Text-to-Image}} & \\
\textbf{Model} & \textbf{MIMIC-CXR} & \textbf{Quilt} & \textbf{DeepEyeNet} & \textbf{MIMIC-CXR} & \textbf{Quilt} & \textbf{DeepEyeNet} & \textbf{AR} \\
\midrule
PMC-OA & 0.139 & 0.142 & 0.152 & 0.152 & 0.149 & 0.157 & 0.148 \\
\dataset & 0.17 & 0.166 & \underline{0.183} & 0.189 & 0.162 & 0.147 & 0.17 \\
BioMedCLIP     & 0.185 & 0.165 & 0.162 & 0.162 & 0.185 & 0.146 & 0.167 \\
BIOMEDICA      & 0.076  & 0.169 & 0.155 & 0.093  & 0.195 & 0.145 & 0.139 \\
\midrule
\datasetcomp   & \textbf{0.25} & \underline{0.203} & 0.172 & \textbf{0.257} & \underline{0.22} & \underline{0.170} & \underline{0.212} \\
\datasetsub    & \underline{0.215} & \textbf{0.226} & \textbf{0.192} & \underline{0.210} & \textbf{0.256} & \textbf{0.197} & \textbf{0.216} \\
\bottomrule
\end{tabular}
}
\label{main_retrieval_results}
\end{table*}

\subsection{Downstream Tasks}
The performance of the encoders is evaluated on several datasets across two primary tasks: retrieval and zero-shot classification. For the retrieval task, we measure both image-to-text and text-to-image retrieval across three datasets representative of distinct medical imaging modalities: Quilt \citep{ikezogwo2024quilt} (microscopy), MIMIC-CXR \citep{johnson2019mimic} (radiology), and DeepEyeNet \citep{huang2021deepopht} (VLP). To evaluate robustness in retrieval, we follow established protocols from \cite{liu2024diffrect} by applying a suite of low-level visual perturbations, including brightness adjustment, spatial shift, rotation, horizontal flip, and zoom, directly to the test images. To assess the statistical significance of robustness differences, we employ the Wilcoxon signed-rank test, a non-parametric method for paired comparisons \citep{wilcoxon1945individual}. We consider a p-value less than 0.01 as statistically significant. For classification, we evaluate models using both zero-shot and linear probing  protocols across a diverse set of tasks: five in radiology, eight in microscopy, and six in VLP. We use our trained vision and text encoders to encode the image and question, respectively.

\subsection{Cross-Modal Retrieval and Robustness}
Table~\ref{main_retrieval_results} and Figure~\ref{fig:intro} (c) (top) summarize the performance of various VLMs on cross-modal retrieval tasks across the three benchmark datasets. We report Recall@200 (other Recall metrics are included in the Appendix), with the final column showing the Average Recall (AR) aggregated across all tasks. Models trained on \datasetsub (subfigures) and \datasetcomp (compound figures) consistently outperform \biomedclip and \biomedica across all tasks and retrieval directions. In particular, \datasetsub sets a new state-of-the-art with an AR of 21.64, showing a 27\% relative improvement in average retrieval performance over the best previous model, \dataset.

\begin{figure}[h]
\centering
\includegraphics[width=0.45\textwidth]{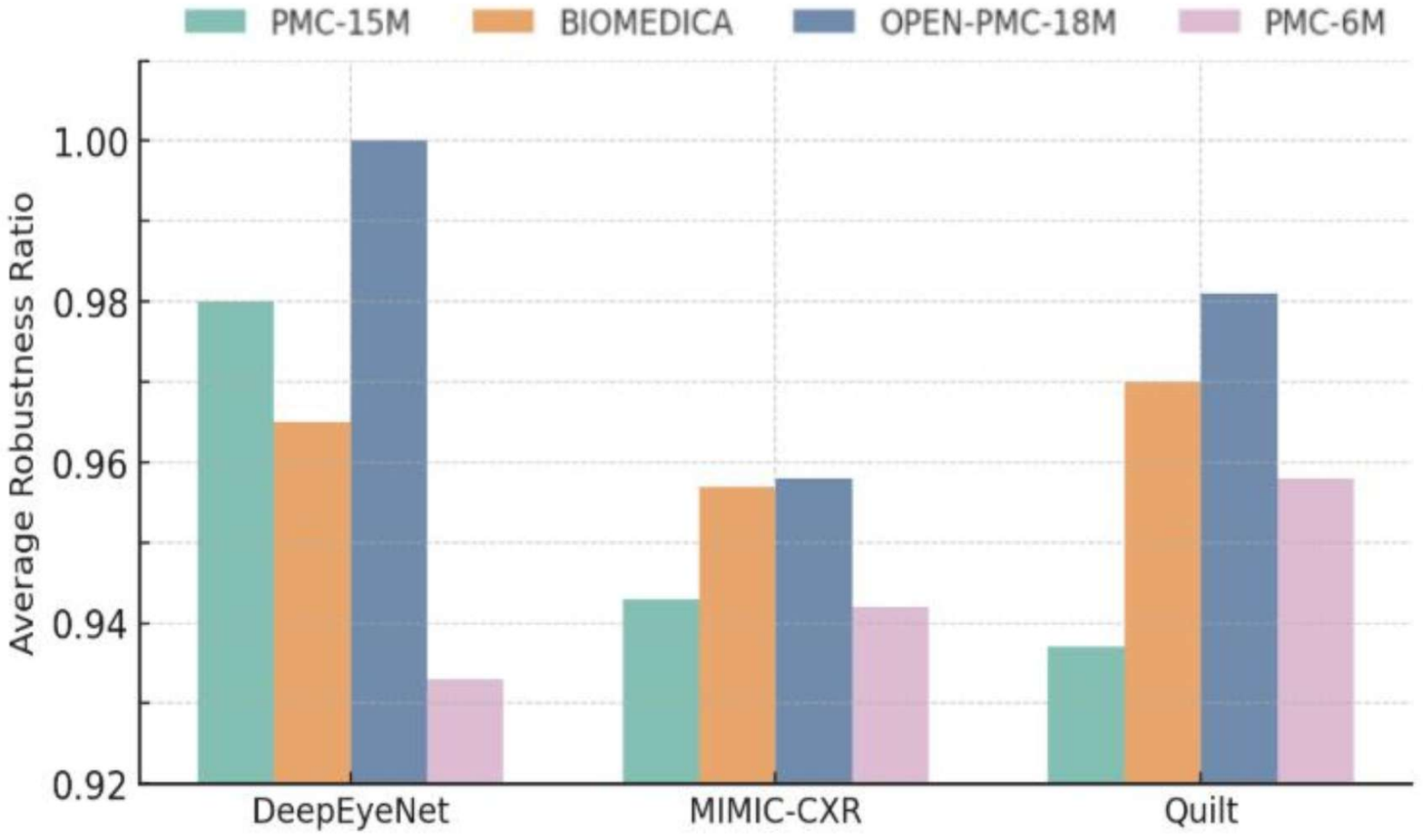}
\caption{Robustness evaluation across retrieval benchmarks. Robustness is quantified as the ratio of performance under visual perturbations (brightness, shift, rotation, flip, zoom) to performance on the original test set. Higher values indicate greater stability to perturbations.}
\label{fig:robust}
\end{figure}

Robustness, quantified as the ratio between retrieval performance under perturbations such as brightness adjustments, shifts, rotations, flips, and 
zoom distortions, and performance on the original data is presented in Figure~\ref{fig:robust}. Models trained on \datasetsub consistently achieve higher robustness scores relative to baseline models, reflecting improved performance stability under input perturbations~\citep{zeng2023pefat} in addition to superior retrieval performances. We observe statistically significant differences ($p < 0.01$) on Quilt and DeepEyeNet. These findings are particularly relevant to our focus on subfigure extraction and the potential for improved robustness in imaging modalities that exhibit high visual and semantic heterogeneity.

\subsection{Zero-shot Classification}
Model comparisons for zero-shot classification are presented in Table~\ref{tab:ZSC_F1_avg}. For each modality, scores are averaged over multiple benchmark datasets (see Appendix Table~\ref{zsc-tab} for details).
Models trained on \datasetsub, achieve the highest overall average, outperforming all existing biomedical VLMs. In particular, \datasetsub shows large gains in microscopy and radiology tasks, and ranks second in radiology demonstrating better transferability relative to all other evaluated models. Across the full set of 19 classification tasks spanning radiology, microscopy, and VLP, \datasetsub ranks first in 13 tasks and second in 1. A similar trend is observed in the linear probing results (Appendix Table~\ref{tab:LP_F1}), where \datasetsub again achieves the highest average performance across modalities.

\begin{table*}[h!]
    \caption{Zero-shot classification average F1-scores across medical modalities.}
    \label{tab:ZSC_F1_avg}
    \centering
    \begin{tabular}{lcccc}
    \toprule
    \textbf{Model} & \textbf{Radiology} & \textbf{VLP} & \textbf{Microscopy} & \textbf{Overall Avg} \\
    \midrule
    PMC-OA      & 30.95 & 29.79 & 42.62 & 34.45 \\
    \dataset    & \textbf{36.19} & 28.30 & 39.12 & 34.54 \\
    BioMedCLIP  & 28.62 & 21.17 & 43.35 & 31.04 \\
    BIOMEDICA   & 29.56 & 26.21 & 45.16 & 33.64 \\
    \midrule
    \datasetcomp & 30.28 & \underline{31.50} & \underline{46.35} & \underline{36.04} \\
    \datasetsub  & \underline{34.55} & \textbf{32.10} & \textbf{55.35} & \textbf{39.77} \\
    \bottomrule
    \end{tabular}
\end{table*}

\subsection{Representation Analysis}
To further understand the effect of dataset scale and curation fidelity on learned representations, we visualize the embedding spaces of models trained on the \datasetcomp (baseline) and \datasetsub datasets.
Both models share identical architectures and training hyperparameters, differing only in dataset composition, allowing a direct comparison of representation quality. We project the embedding spaces of three benchmark sets, each constructed by combining datasets used for retrieval and zero-shot classification across radiology, microscopy, and VLP, into two dimensions using t-SNE (Figure~\ref{fig:tsne} in the Appendix). The radiology benchmark includes MIMIC-CXR and other related zero-shot classification tasks, totaling approximately 41,000 samples. The microscopy and VLP benchmarks contain approximately 20,000 and 6,000 samples, respectively. To quantify differences between the embedding distributions, we compute the Maximum Mean Discrepancy (MMD) \cite{gretton2012kernel} (more details in \S\ref{RepAnal}). 

Visual inspection of the embeddings reveals distinct representational structures in the latent spaces of the models. Moreover, the MMD analysis confirms that the observed differences are statistically significant across all modalities. These results confirm that models trained on subfigure-level data with enriched textual context learn substantially different representations compared to models trained on compound figures and full captions without summaries.

\iffalse
\begin{figure*}[h]
    \centering
    \includegraphics[width=1.0\textwidth]{figure/new-tsne.pdf}
    \caption{t-SNE visualizations of models embeddings trained on \datasetsub and \datasetcomp on three imaging modalities, illustrating the structure and separation of the learned representation spaces. MMD analysis reveals statistically significant differences in embedding distributions across all imaging modalities. }
    \label{fig:tsne}
\end{figure*}
\fi

\section{Effects of Fine-Grained and Context-Enriched Language Supervision on Medical Representation Learning}
\label{sec:ablation}
We perform an ablation to examine the role of increasing textual fidelity. The results of this analysis are summarized in Figure~\ref{fig:intro} (c) (bottom). To make the experiment computationally feasible, we restrict training to the radiology subset of \datasetsub, which contains 3.2 million image–text pairs. All model variants use the same encoder architecture, optimization settings, and contrastive pretraining objective. 

In the \textit{Subfigure-only} variant, each subfigure is paired with its full caption. In \textit{Subfigure + Subcaption}, the full caption is replaced by the extracted subcaption corresponding to that subfigure, providing finer-grained text alignment. Finally, in \textit{Subfigure + Subcaption + Summary}, each subcaption is further augmented with a summary of its in-text context, capturing additional diagnostic or experimental details.

Results in Figure~\ref{fig:intro}(c) (bottom) on MIMIC-CXR show that replacing full captions with subcaptions alone does not yield a performance gain, in fact, the reduced textual richness leads to a slight drop in average recall, highlighting that subcaptions, while more locally precise, may be too sparse to support robust alignment. However, when contextual summaries are added, performance increases substantially, indicating that the combination of panel-specific subcaptions and enriched contextual information provides the strongest supervision signal. Together, these findings suggest that high-quality biomedical representations rely not only on correcting local misalignment through subcaptions, but also on restoring broader semantic and clinical grounding through contextual summaries.

\section{Limitations and Challenges}
\label{sec:limit}

Our findings suggest that in the context of vision-language representation learning, data quality and scale should be viewed as complementary axes in building effective and robust biomedical VLMs. Subfigure extraction, used here as a means to improve alignment quality, demonstrates clear benefits, particularly in visually heterogeneous domains such as microscopy and visible light photography, as shown in Figures~\ref{fig:intro} and~\ref{fig:layout}. While our results highlight promising trends, additional analysis is required to fully assess generalization.
Beyond representation quality, it is important to explore the integration of our encoders with large language model decoders for downstream tasks that involve generative reasoning over visual inputs, such as medical report generation and visual question answering, to establish a more grounded framework for multimodal clinical applications.
Future work should also evaluate the factual consistency of these encoder, decoder systems relative to existing baselines, to determine whether well-curated, high-fidelity supervision improves not only discriminative performance but also the reliability of downstream generative tasks.

We recognize that scaling and curating large biomedical datasets brings challenges that extend beyond improving model performance. To support transparency and reproducibility, we release the dataset, data curation code, subfigure detection models, and training pipelines. However, interpretability remains an open challenge in VLMs and particularly in the biomedical domain. Although our models are not intended for clinical deployment, they could be fine-tuned or adapted for various clinical applications. However, without rigorous validation and careful consideration of clinical safety, such use poses serious risks. 

Furthermore, our datasets, sourced from open-access repositories such as PubMed Central, may reflect underlying biases tied to specific institutions, imaging protocols, or publication norms. These factors can influence model behavior in subtle ways, limiting generalizability, especially when applied to underrepresented populations or distinct clinical settings. 

\section{Related Work}
\label{sec:related}

\subsection{Biomedical Vision-Language Datasets}
Most efforts to date have relied on mining figures and captions from the PMC Open Access subset.\footnote{https://pmc.ncbi.nlm.nih.gov/tools/openftlist/} One of the earliest publicly available datasets is ROCO \citep{pelka2018radiology}, which compiled around 80,000 radiology and 6,000 non-radiology images, enriched with metadata such as captions and keywords. Later, \citet{lin2023pmc} introduced PMC-OA , which includes 1.6 million image-text pairs. Their contribution emphasized automation, proposing a pipeline to streamline the pairing process and reduce human annotation. More recently, \citet{zhang2023biomedclip} announced \biomedclip, a dataset of 15 million image-text pairs. The largest released dataset to date is \biomedica \citep{lozano2025biomedica}, which comprises 24 million pairs and employs clustering, vision encoders, and expert labeling to assign modalities at global and local levels. While these efforts represent major progress at scale, recent work has emphasized that data quality is a critical factor in learning effective and generalizable medical representations \citep{openpmc}. Building on the premise of \datasetsmall, our work takes a quality-first approach while also significantly scaling up the dataset. %\arash{The rest of this paragraph should be moved out of Related work.}We expand the corpus nearly nine-fold, constructing a high-quality biomedical image-text dataset with broader coverage across imaging modalities and clinical subdomains. In parallel, we improve subfigure segmentation by training a multimodal panel detector that extends prior work with enhanced support for microscopy images—a modality underrepresented or misclassified in earlier segmentation efforts. This detector is trained on a large synthetic corpus and applied to compound figures across PMC, enabling robust and scalable subfigure-level pairing that preserves both visual and semantic granularity.

\subsection{Subfigure Extraction as Object Detection}
Early approaches to compound figure separation relied on classical computer vision techniques, using heuristics based on whitespace, edge detection, or layout regularity. However, these methods often struggled to handle diverse panel styles and complex spatial arrangements. More recent work treats subfigure extraction as an object detection problem, leveraging deep learning models. For example, \citet{tsutsui2017data} and \citet{yao2021compound} used YOLO for subfigure separation. \citet{lin2023pmc} also uses an object detection model to extract subfigures in their pipeline. They train a DETR (DEtection TRansformer) model \citep{carion2020end} on the MedICaT dataset \citep{subramanian2020medicat} containing 2069 annotated compound figures.

Data annotation for training an image decomposition model is challenging and time-consuming. Current annotated datasets for this purpose are small, which lead to models with suboptimal performance. To overcome this, synthetic datasets of compound figures have been proposed, where subfigures are programmatically composed to simulate real-world layouts. This allows training of object detection models without relying on large-scale human-annotated data \citep{tsutsui2017data, yao2021compound}. %\arash{The rest of the paragraph should be moved out of Related Work} Our work builds on this line of research by generating a large-scale synthetic dataset of 500K compound figures for training a high-performance DAB-DETR model. We demonstrate that this detector generalizes well to real PMC figures and enables scalable subfigure decomposition, yielding over 18M image-caption pairs.

\section{Conclusion}
In this work, we addressed a fundamental limitation in current biomedical vision–language pretraining: the reliance on misaligned compound figures and weak medical context in image-text pairs. We introduced \datasetsub, a large-scale, high-fidelity dataset constructed through accurate subfigure extraction and contextual text enrichment. Our systematic evaluations demonstrate that models trained on \datasetsub achieve consistent gains across radiology, microscopy, and visible-light photography, outperforming existing PMC-derived datasets in retrieval, classification, and robustness. Beyond empirical improvements, our findings highlight a broader insight: effective biomedical representation learning depends more on precise data curation and alignment than on scale alone. We believe this dataset, and the accompanying analysis, provides a principled path toward multimodal models that are better aligned with the heterogeneous and context-dependent nature of real biomedical data.
%In this paper we addressed a critical gap in the design of high-fidelity multimodal medical datasets, aiming to advance robust and generalizable representation learning. We evaluated the effectiveness and robustness of subfigure extraction. We introduced \datasetsub, one of the largest and highest quality image-caption pairs to date. Models trained on \datasetsub consistently outperform existing benchmarks across radiology, microscopy, and visible light photography. These findings lay the groundwork for more generalizable medical VLMs and better aligned with complex realities of biomedical data.

%\input{CVPR/sec/0_abstract}    
%\input{CVPR/sec/1_intro}
%\input{CVPR/sec/2_related}
%\input{CVPR/sec/3_data}
%\input{CVPR/sec/4_experiments}
%\input{CVPR/sec/5_ablation}
%\input{CVPR/sec/6_limitation}
%\input{CVPR/sec/7_conclusion}
{
    \small
    \bibliographystyle{ieeenat_fullname}
    \bibliography{main}
}

\appendix

\clearpage
\setcounter{page}{1}
\maketitlesupplementary

\section{Additional Dataset Statistics}
\label{sec:data_stats_supp}
\begin{figure*}[t]
    \centering
    % Row 1
    \begin{subfigure}[t]{0.48\textwidth}
        \centering
        \includegraphics[width=\linewidth]{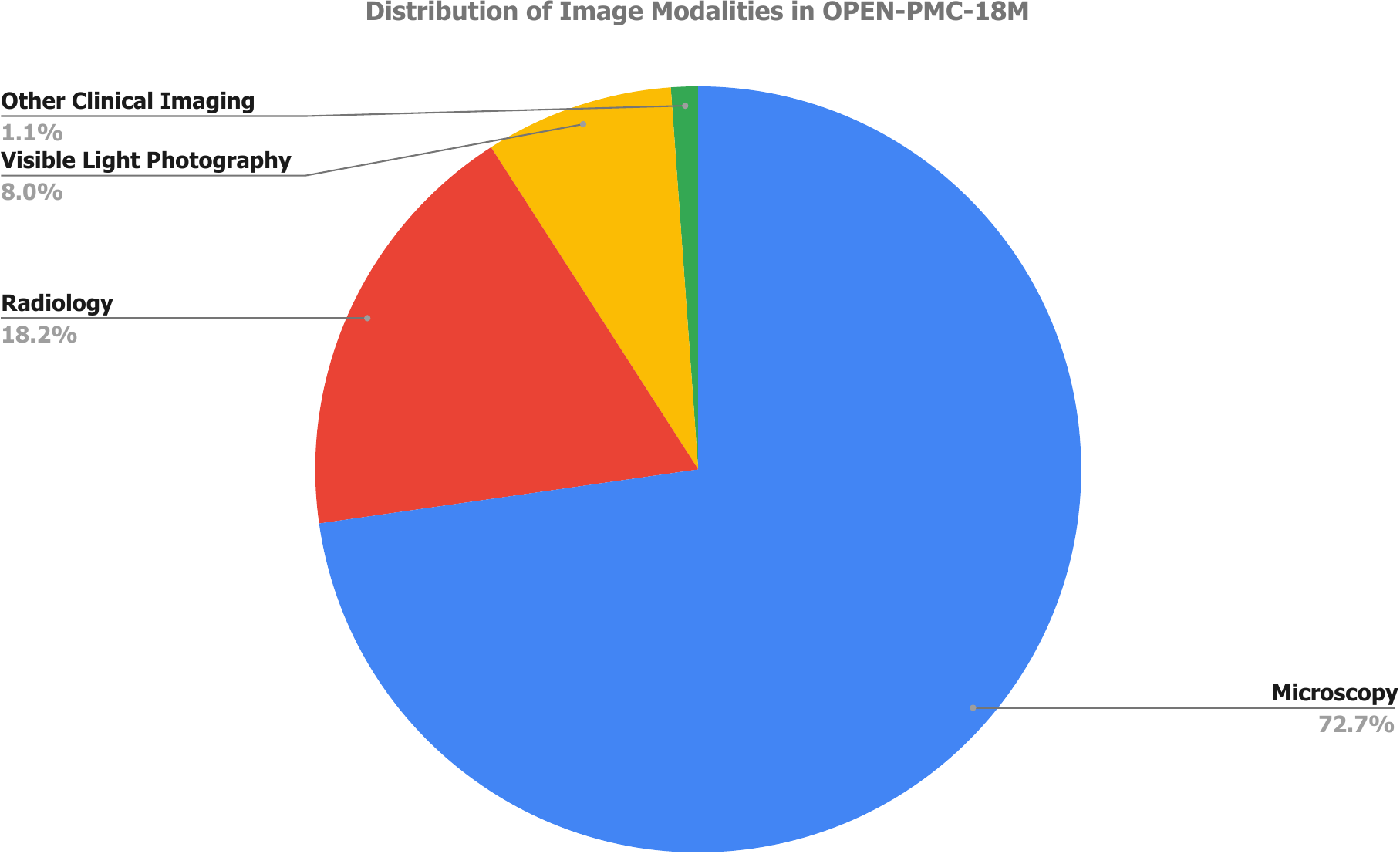}
        \caption{Distribution of image modalities in \datasetsub.}
        \label{fig:modality}
    \end{subfigure}
    \hfill
    \begin{subfigure}[t]{0.48\textwidth}
        \centering
        \includegraphics[width=\linewidth]{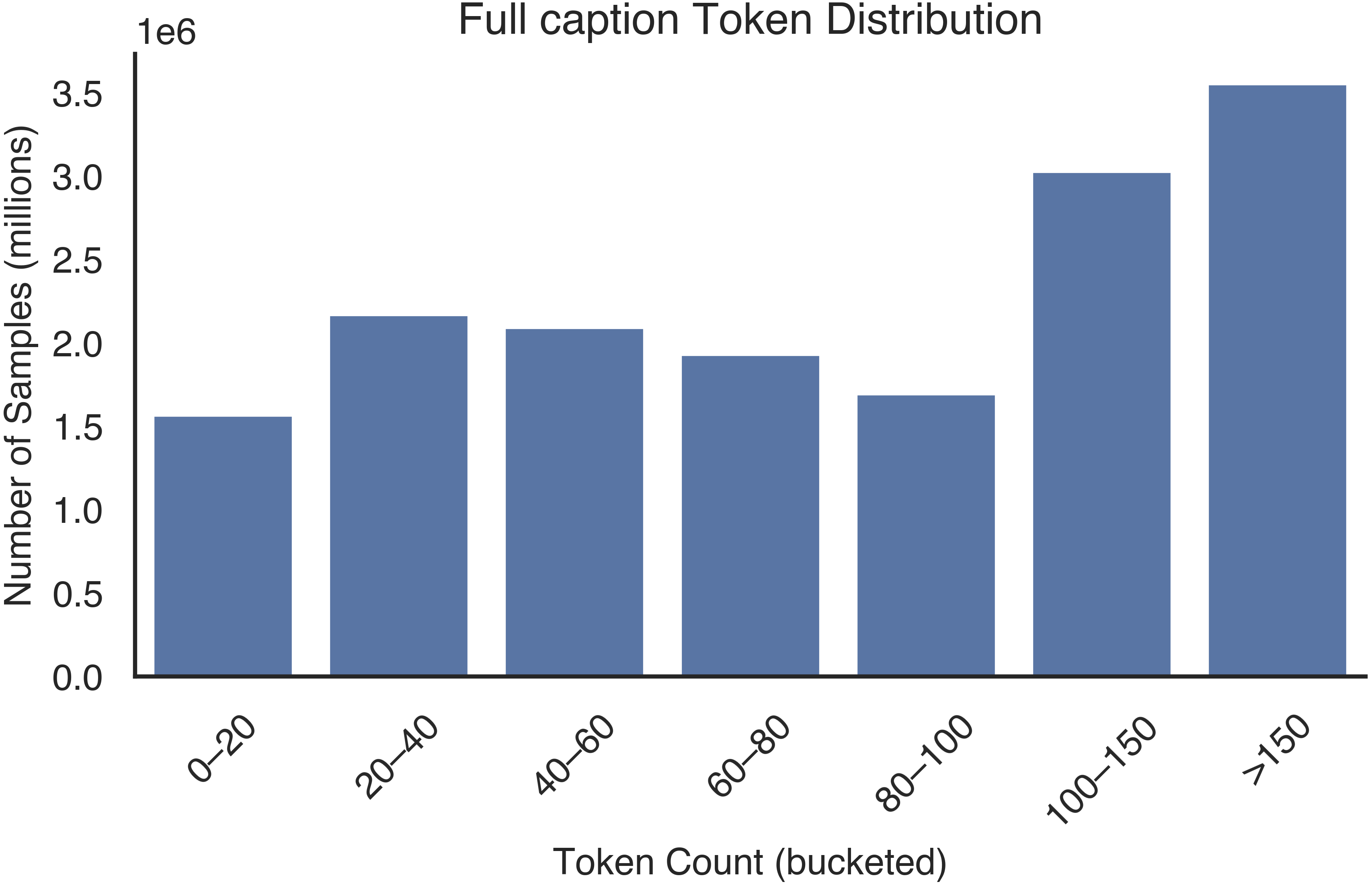}
        \caption{Full caption token distribution.}
        \label{fig:caption_dist}
    \end{subfigure}
    
    % Row 2
    \vspace{2mm}
    \begin{subfigure}[t]{0.48\textwidth}
        \centering
        \includegraphics[width=\linewidth]{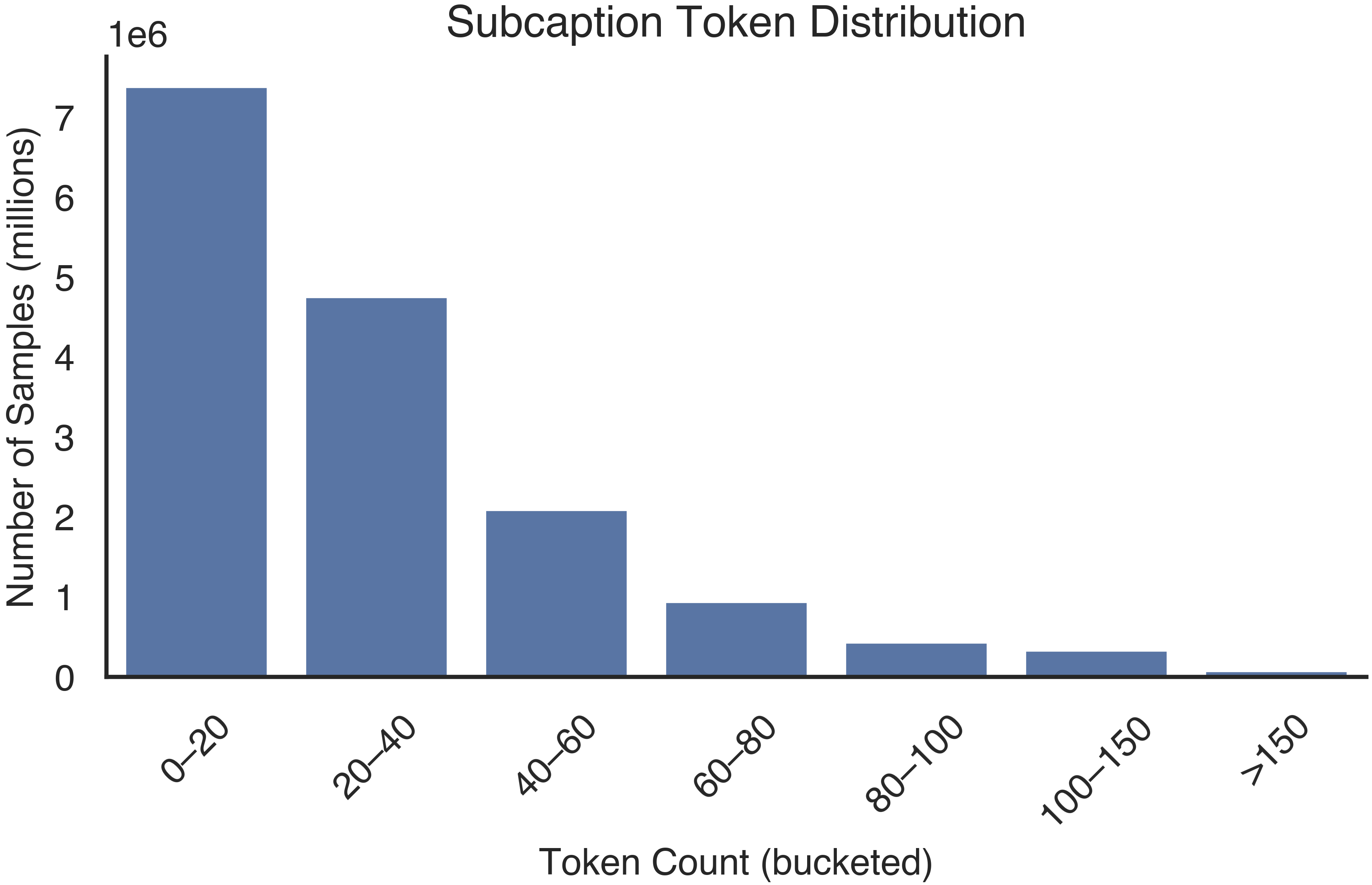}
        \caption{Subcaption token distribution.}
        \label{fig:subcaption_dist}
    \end{subfigure}
    \hfill
    \begin{subfigure}[t]{0.48\textwidth}
        \centering
        \includegraphics[width=\linewidth]{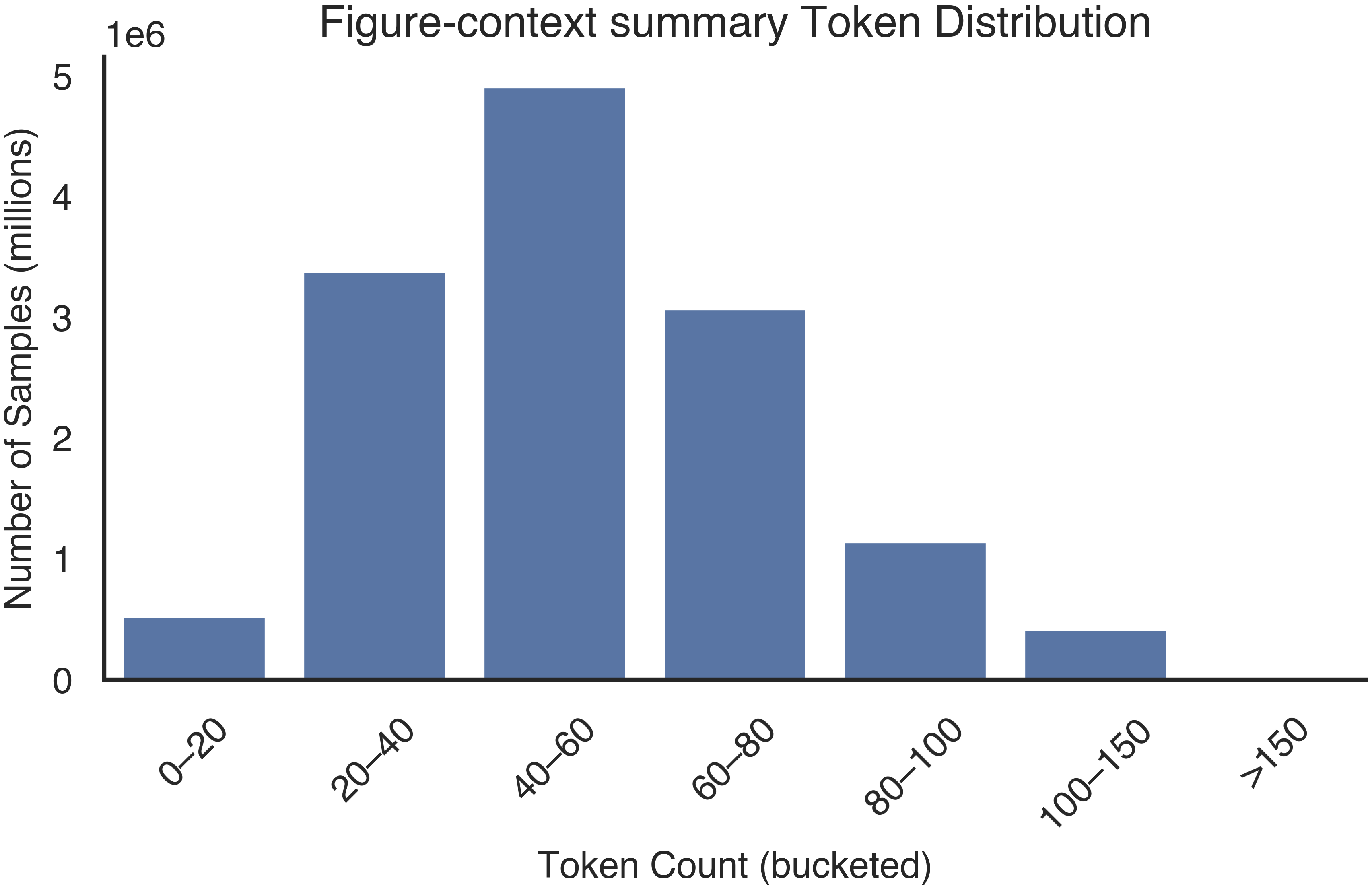}
        \caption{Figure-context summary token distribution.}
        \label{fig:summary_dist}
    \end{subfigure}

    \caption{
    Dataset composition and text length statistics.
    (a) Distribution of image modalities in \datasetsub.
    (b) Token distribution for full captions.
    (c) Token distribution for subcaptions.
    (d) Token distribution for figure-context summaries.
    }
    \label{fig:supp_dataset_statistics}
\end{figure*}

We provide additional dataset statistics in visualizations in this section (see Figure \ref{fig:supp_dataset_statistics}).

\section{Training Details}
\label{sec:train_det_supp}
\subsection {Encoder Pretraining}
\label{subsec:enc-pretrain}
As a first step, we train separate encoders for image and text modalities by aligning their representations using a vanilla contrastive loss. Let $\ienc$ denote an image encoder and $\tenc$ denote a text encoder that maps images and text to a common representation space, respectively. Given a batch of training samples $B = \{ (x_i, t_i)\}_{i=1}^N$, where $x_i$ and $t_i$ denote the $i^{\text{th}}$ image and text instances respectively, the InfoNCE loss \citep{oord2018representation} is optimized by minimizing the distance between the representations of an image and its corresponding text, $(\ienc(x_i), \tenc(t_i))$, while maximizing the distance between unrelated image-text representation pairs, $(\ienc(x_i), \tenc(t_j)), \hspace{1mm} i \neq j$:
% \beqa
% \label{eq:contrastive_loss}
% \ell_{\text{con}}(x_i,t_i;B) = - \left( \log \frac{\exp(\inl \ienc(x_i), \tenc(t_i) \inr / \tau)}{\sum_{k=1}^N \exp (\inl \ienc(x_i), \tenc(\boldsymbol{t_k}) \inr / \tau)} +
%  \log \frac{\exp(\inl \ienc(x_i), \tenc(t_i) \inr / \tau)}{\sum_{k=1}^N \exp (\inl \ienc(\boldsymbol{x_k}), \tenc(t_i) \inr / \tau)} \right),
% \eeqa
% \saidul{I think we used the same contrastive loss equation previously in section 3}
% \arash{Let's move the contrastive loss formula to the appendix. It's well-known and there's little value in including it.}
\beqa
\label{eq:contrastive_loss}
\begin{aligned}
\ell_{\text{con}}(x_i,t_i;B) = - \Bigg(
    &\log 
      \frac{\exp(\inl \ienc(x_i), \tenc(t_i) \inr / \tau)}
           {\smashoperator{\sum_{k=1}^N} \exp(\inl \ienc(x_i), \tenc(\mathbf{t_k}) \inr / \tau)}
      \\
    &+
      \log 
      \frac{\exp(\inl \ienc(x_i), \tenc(t_i) \inr / \tau)}
           {\smashoperator{\sum_{k=1}^N} \exp(\inl \ienc(\mathbf{x_k}), \tenc(t_i) \inr / \tau)}
\Bigg)
\end{aligned}
\eeqa
where $\inl \cdot, \cdot \inr$ denotes similarity between two vectors (e.g. cosine similarity), and $\tau > 0$ is a temperature parameter. For simplicity of notation, we drop $B$ and denote the loss for $(x,t)$ by $\loss_\text{con}(x,t)$. Multimodal contrastive learning trains encoders $\ienc$ and $\tenc$ by minimizing Eq.~\ref{eq:contrastive_loss} over the pairs in $B$:
\beqa
\label{eq:multimodal_loss}
\loss_{\text{multimodal}} = \min_{\ienc, \tenc} \hspace{2mm} \mathbb{E}_B \Big[ \frac{1}{N} \sum_{i=1}^N  \ell_\text{con}(x_i,t_i)\Big].
\eeqa

% \subsection{Pretraining Hyperparameters}
\subsection{Training Hyperparameters}

For pretraining, we use the AdamW optimizer with a weight decay of $0.2$, $\beta_1 = 0.9$, and $\beta_2 = 0.98$. The learning rate is scheduled using cosine decay, with a linear warmup over the first $10\%$ of the total training steps. We apply gradient accumulation with a frequency of 4. Training is performed on 8 NVIDIA A100 GPUs with a total batch size of 2048. The initial learning rate is set to $5.0 \times 10^{-4}$, and models are trained for 64 epochs.

\subsection{Performance on Validation set}
\label{val}
\begin{figure}[h!]
    \centering
    \includegraphics[width=0.9\linewidth]{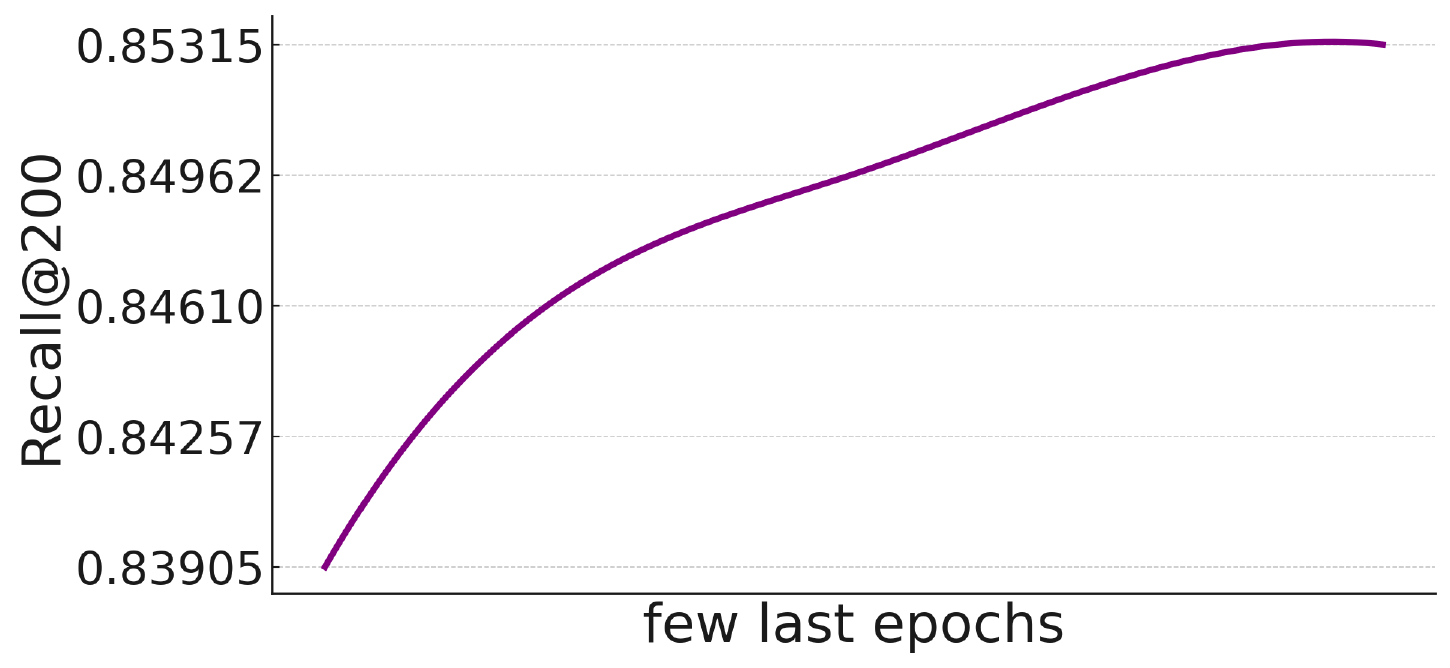}
    \caption{Recall@200 for the last epochs of the trained model on the validation data}
    \label{fig:valid}
\end{figure}

We use a 50,000-sample subset of the dataset as our validation set. Each sample includes a sub-figure and its paired sub-caption. Retrieval is used as the validation task, and recall@200 is the metric for selecting the best epoch. Figure~\ref{fig:valid} presents the recall@200 scores for the final epochs of training. The results show a rising trend that eventually levels off, indicating stable validation performance.

\section{Evaluation Datasets}
We evaluate our pretrained models on three downstream tasks: image-text retrieval, zero-shot classification, and linear probing. A summary of the evaluation datasets used for these tasks is provided in Table~\ref{tab:eval-data}.

\begin{table*}[t]
\small
  \caption{Evaluation datasets.}
  \label{tab:eval-data}
  \centering
  \resizebox{\textwidth}{!}{
  \begin{tabular}{m{2cm} m{2cm} m{3cm} m{3.5cm} c}
    \toprule
    \textbf{Task} & \textbf{Setup} & \textbf{Dataset} & \textbf{Modality} & \textbf{Nb. Samples}  \\
    \midrule
    \multirow{4}{2cm}{Retrieval} & \multirow{4}{2cm}{\itt ~\& \tti} & Quilt-1M & Histopathology & 13,559 \\
    &  & MIMIC-IV-CXR & Chest X-ray & 3,269  \\
    &  & DeepEyeNet & Retina & 3,140 \\
    \midrule
    \multirow{22}{2cm}{ Zero-shot classification \& Linear probing} 
    & 6 classes & PAD-UFES-20 & Dermatology & 460  \\
    & 7 classes & SkinCancer & Dermatology & 2,003  \\
    & 2 classes & PatchCamelyon (PCam) & Histopathology & 32,768  \\
    & 8 classes & NCT-CRC-HE-100K & Histopathology & 6,333  \\
    & 3 classes & LC25000Lung & Histopathology & 3,000  \\
    & 2 classes & LC25000Colon & Histopathology & 2,000 \\
    & 4 classes & BACH & Histopathology & 100 \\
    & 4 classes & SICAPv2 & Histopathology & 2,122  \\
    & 9 classes & PathMNIST+ & Colon Pathology & 107,180  \\
    & 7 classes & DermaMNIST+ & Dermatoscope & 10,015  \\
    & 4 classes & OctMNIST+ & Retinal OCT & 109,309  \\
    & 2 classes  & PneumoniaMNIST+ & Chest X-Ray & 5,856  \\
    & 5 classes & RetinaMNIST+ & Fundus Camera & 1,600 \\
    & 2 classes  & BreastMNIST+ & Breast Ultrasound & 780  \\
    & 8 classes & BloodMNIST+ & Blood Cell Microscope & 17,092  \\
    & 8 classes & TissueMNIST+ & Kidney Cortex Microscope & 236,386  \\
    & 11 classes & OrganAMNIST+ & Abdominal CT & 58,830  \\
    & 11 classes & OrganCMNIST+ & Abdominal CT & 23,583  \\
    & 11 classes & OrganSMNIST+ & Abdominal CT & 25,211 \\
    \bottomrule
  \end{tabular}%
  }
\end{table*}

\paragraph{MIMIC-CXR:} MIMIC-CXR contains 377,110 de-identified chest X-ray images from 65,379 patients, accompanied by free-text reports. The dataset was collected from the emergency department of Beth Israel Deaconess Medical Center. Each patient typically has multiple views and a corresponding radiology report labeled using the CheXpert labeling tool (Irvin et al., 2019), which identifies 13 common conditions such as atelectasis, cardiomegaly, consolidation, pleural effusion, and pneumonia.

\paragraph{Quilt-1M:} Quilt-1M comprises over one million histopathology image-text pairs. The largest subset, Quilt, includes 802,144 pairs extracted from 1,087 hours of educational histopathology videos on YouTube. Captions were generated using a combination of large language models, handcrafted rules, and automatic speech recognition. Additional subsets come from PubMed Open Access, LAION-5B, and OpenPath Twitter data, resulting in a combined dataset of over one million pairs.

\paragraph{DeepEyeNet:} DeepEyeNet is a large-scale retinal image dataset comprising 15,709 images, including both color fundus photography (CFP) and fluorescein angiography (FA). Each image is annotated with three expert-defined labels: a disease or symptom name, a set of relevant keywords, and a detailed clinical description. The dataset covers 265 distinct retinal conditions.

\paragraph{SICAP:} The Prostate Cancer Grade Assessment (SICAP) dataset comprises prostate histology whole-slide images annotated with global Gleason scores and path-level Gleason grades, supporting research in automated prostate cancer grading.

\paragraph{PAD-UFES-20:} PAD-UFES-20 includes 2,298 clinical images of six types of skin lesions, each accompanied by up to 22 patient metadata features, facilitating studies in skin lesion classification.

\paragraph{Skin Cancer:} This dataset contains 2,357 dermatoscopic images of skin lesions, labeled with diagnostic categories, aiding in the development of skin cancer detection models.

\paragraph{PCam (PatchCamelyon):} PCam consists of 327,680 color images (96×96 px) extracted from histopathologic scans of lymph node sections, each labeled to indicate the presence of metastatic tissue.

\paragraph{NCT-CRC-HE:} The NCT-CRC-HE dataset comprises 100,000 non-overlapping image patches from H\&E-stained histological images of human colorectal cancer and normal tissue, supporting research in histopathological image analysis.

\paragraph{LC-Lung:} LC-Lung includes 15,000 histopathological images of lung tissue, categorized into benign and malignant classes, useful for lung cancer classification studies.

\paragraph{LC-Colon:} LC-Colon comprises 10,000 histopathological images of colon tissue, labeled as benign or malignant, aiding in colon cancer detection research.

\paragraph{BACH:} The Breast Cancer Histology (BACH) dataset contains microscopy images of breast tissue, annotated across four classes: normal, benign, in situ carcinoma, and invasive carcinoma, facilitating automated breast cancer diagnosis.

\paragraph{DermaMNIST+:} DermaMNIST+ consists of 10,015 dermatoscopic images categorized into seven skin disease classes, serving as a benchmark for skin lesion classification tasks.

\paragraph{OCTMNIST+:} OCTMNIST+ includes 109,309 optical coherence tomography images labeled for retinal diseases like choroidal neovascularization, diabetic macular edema, and drusen, supporting ophthalmic image classification.

\paragraph{PneumoniaMNIST+:} PneumoniaMNIST+ is based on 5,856 pediatric chest X-ray images, labeled for pneumonia detection, aiding in the development of automated pneumonia diagnosis models.

\paragraph{RetinaMNIST+:} RetinaMNIST+ comprises 1,600 retinal fundus images labeled for common eye diseases, useful for training models in automated retinal disease classification.

\paragraph{BreastMNIST+:} BreastMNIST+ contains 780 ultrasound images of breast tumors, labeled as benign or malignant, supporting breast cancer detection research.

\paragraph{BloodMNIST+:} BloodMNIST+ consists of 17,092 microscopic images of blood cells, classified into eight cell types, facilitating automated classification tasks in hematology.

\paragraph{TissueMNIST+:} TissueMNIST+ includes 236,386 microscopic images of tissue samples from different organs, labeled according to tissue type, supporting histopathological analysis.

\paragraph{PathMNIST+:} PathMNIST+ is derived from colorectal cancer tissue slides, containing 107,180 images labeled with nine different tissue classes, aiding in multi-class classification tasks in pathology.

\paragraph{OrganAMNIST+:} OrganAMNIST+ consists of 58,850 abdominal CT images labeled with different anatomical organ classes, supporting organ segmentation and classification tasks.

\paragraph{OrganCMNIST+:} OrganCMNIST+ contains 23,600 coronal CT images of various organs, labeled for organ classification tasks, used for research in medical image understanding.

\paragraph{OrganSMNIST+:} OrganSMNIST+ comprises 23,600 sagittal CT images of multiple organs, annotated for classification, aiding in comprehensive medical imaging analysis.

\begin{table*}[t]
\centering
\caption{Retrieval performance (Recall@10) of all models trained on paired image-caption pairs in the medical domain. The last column, Average Recall (AR), aggregates the results across all tasks.}
{\small
\resizebox{\textwidth}{!}{
\begin{tabular}{lccc|ccc|c}
\toprule
              & \multicolumn{3}{c|}{\textbf{Image-to-Text}} & \multicolumn{3}{c|}{\textbf{Text-to-Image}} & \\
\textbf{Model} & \textbf{MIMIC} & \textbf{Quilt} & \textbf{DeepEyeNet} & \textbf{MIMIC} & \textbf{Quilt} & \textbf{DeepEyeNet} & \textbf{AR} \\
\midrule
PMC-OA         & 0.014 & 0.020 & 0.026 & 0.010 & 0.016 & 0.017 & 0.017 \\
\dataset       & 0.022 & 0.018 & 0.024 & 0.016 & 0.016 & 0.024 & 0.020 \\
BioMedCLIP     & 0.022 & 0.024 & 0.031 & 0.015 & 0.027 & 0.024 & 0.023 \\
BIOMEDICA      & 0.005 & 0.033 & 0.023 & 0.006  & 0.041 & 0.022 & 0.021 \\
\midrule
\datasetcomp   & 0.033 & 0.028 & 0.039 & 0.028 & 0.032 & 0.035 & 0.032 \\
\datasetsub    & 0.023 & 0.033 & 0.033 & 0.019 & 0.039 & 0.041 & 0.031 \\
\bottomrule
\end{tabular}%
}
}
\label{ret-table-10}
\end{table*}
\begin{table*}[h]
\centering
\caption{Retrieval performance (Recall@50) of all models trained on paired image-caption pairs in the medical domain. The last column, Average Recall (AR), aggregates the results across all tasks.}
{\small
\resizebox{\textwidth}{!}{
\begin{tabular}{lccc|ccc|c}
\toprule
              & \multicolumn{3}{c|}{\textbf{Image-to-Text}} & \multicolumn{3}{c|}{\textbf{Text-to-Image}} & \\
\textbf{Model} & \textbf{MIMIC} & \textbf{Quilt} & \textbf{DeepEyeNet} & \textbf{MIMIC} & \textbf{Quilt} & \textbf{DeepEyeNet} & \textbf{AR} \\
\midrule
PMC-OA         & 0.054 & 0.062 & 0.071 & 0.044 & 0.056 & 0.070 & 0.059 \\
\dataset       & 0.072 & 0.059 & 0.058 & 0.056 & 0.053 & 0.077 & 0.062  \\
BioMedCLIP     & 0.067 & 0.070 & 0.074 & 0.055 & 0.082 & 0.074 & 0.070  \\
BIOMEDICA      & 0.024  & 0.084 & 0.071 & 0.030  & 0.102 & 0.067 & 0.63 \\
\midrule
\datasetcomp   & 0.106 & 0.087 & 0.092 & 0.097 & 0.098 & 0.088 & 0.094 \\
\datasetsub    & 0.078 & 0.105 & 0.090 & 0.072 & 0.113 & 0.105 &  0.093 \\
\bottomrule
\end{tabular}%
}
}
\label{ret-table-50}
\end{table*}
\subsection{Representations Analysis}
\label{RepAnal}
To explore differences in the structure of learned image representations, we project the embedding spaces of three benchmark sets, each constructed by combining datasets used for retrieval and zero-shot classification across radiology, microscopy, and visible light photography (VLP), into two dimensions using t-SNE (Figure~\ref{fig:tsne}). The radiology benchmark includes MIMIC-CXR and other related zero-shot classification tasks, totaling approximately 41,000 samples. The microscopy and VLP benchmarks contain approximately 20,000 and 6,000 samples, respectively. To quantify differences between the embedding distributions, we compute the Maximum Mean Discrepancy (MMD) \cite{gretton2012kernel}. Given a dataset $X$ (e.g., all radiology samples), we extract embeddings $\phi(X)$ and $\psi(X)$ using vision encoders $\phi$ and $\psi$ trained on \datasetsub and \datasetcomp, respectively. To assess whether the differences between these distributions are statistically significant, we perform a permutation test by randomly reassigning samples and recomputing MMD over 100 iterations to generate an empirical null distribution.

Visual inspection of the embeddings reveals distinct representational structures between the two models. This distinction is particularly evident in microscopy and VLP, where the latent spaces of the two models are more clearly differentiated. In contrast, radiology embeddings appear more intermixed, with less visual separation between the models’ representation spaces. Nonetheless, the MMD analysis confirms that the observed differences are statistically significant across all modalities. For the aggregated radiology dataset, the observed MMD is 0.0214 (null range: 0.0186–0.0214; \textit{p} = 0.005). For the aggregated microscopy dataset, the observed MMD is 0.0212 (null range: 0.0188–0.0212; \textit{p} $<$ 0.001). For the VLP dataset, the observed MMD is again 0.0214 (null range: 0.0186–0.0214; \textit{p} = 0.007). These results indicate that models trained on subfigure-level data yield significantly different representation spaces compared to those trained on compound figures.

\begin{figure*}[h]
    \centering
    \includegraphics[width=1.0\textwidth]{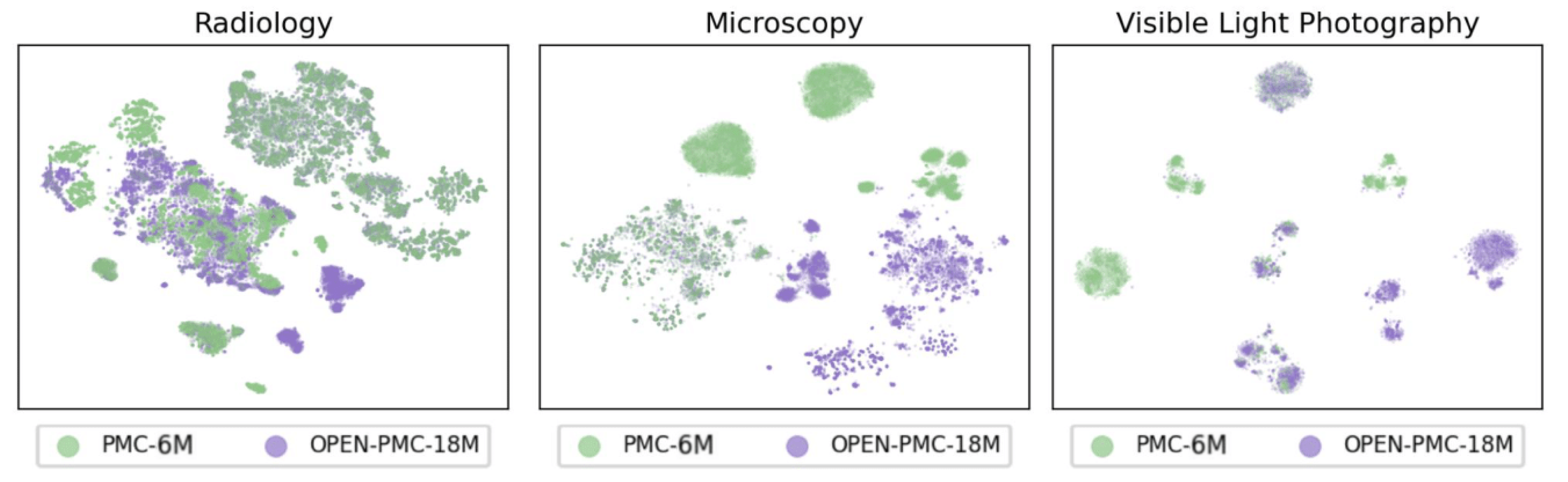}
    \caption{t-SNE visualizations of models embeddings trained on \datasetsub and \datasetcomp on three imaging modalities, illustrating the structure and separation of the learned representation spaces. MMD analysis reveals statistically significant differences in embedding distributions across all imaging modalities. }
    \label{fig:tsne}
\end{figure*}

\section{Prompts}
\label{sec:prompts}
We present our subcaption extraction, summary generation and subcaption extraction evaluation judge prompt in this section.
Figure~\ref{fig:subcap} illustrates the prompt used for subcaption extraction with the Qwen2.5-VL-32B-Instruct model. Figure~\ref{fig:summary} presents the prompt for summary generation using the Qwen2.5-14B-Instruct model. 
% Figure~\ref{fig:subcap-judge} shows the prompt used for evaluating sub-caption correctness with the Qwen2.5-VL-32B-Instruct judge model.

\begin{figure*}[t]
% \begin{tcolorbox}[
%     enhanced,
%     breakable,
%     colback=black!80,
%     coltext=white,
%     left=2mm, right=2mm, top=1mm, bottom=1mm,
%     boxrule=0pt,
%     sharp corners
% ]
% \textbf{\large Prompt: Sub-caption extraction}
% \end{tcolorbox}

% \begin{tcolorbox}[
%     enhanced,
%     breakable,
%     colback=gray!10,
%     colframe=gray!50,
%     left=3mm, right=3mm, top=2mm, bottom=2mm,
%     boxrule=0.3pt,
%     sharp corners
% ]
\begin{tcolorbox}[
    enhanced,
    % breakable,
    colback=gray!10,
    colframe=gray!50,
    colbacktitle=black!80,
    coltitle=white,
    title=\large \textbf{Prompt: Subcaption extraction},
    left=3mm, right=3mm, top=1.5mm, bottom=1.5mm,
    boxrule=0.3pt,
    sharp corners,
]
\ttfamily
INSTRUCTIONS:\\
You are an expert medical image image captioning assistant. Your task is the following:\\
1. You will be provided with a subfigure image that is part of a full image figure and the full figure caption in the input.\\
2. The full caption contains descriptions for multiple subfigures (e.g., Subfigure-A, Subfigure-B, etc.).\\
3. Your task is to identify the relevant subfigure caption corresponding to the provided subfigure image from the full caption exactly as it appears.\\
4. If the subcaption is written jointly for two or more subfigures (e.g., A-C together, (A-C), Axial (A) and coronal (B), etc.), copy that combined description exactly as it appears.\\
5. Do NOT rewrite, summarize, or generate new text. Copy the relevant portion exactly as it appears in the full caption.\\
6. Here, "exactly as it appears" means the extracted caption must match word-for-word, character-for-character with the correct subfigure caption text from the full caption. It must be a verbatim copy, not paraphrased, summarized, or partially copied.\\
7. If no relevant caption is found in the full caption, output the verbatim copy of the entire full caption.\\[1mm]
OUTPUT FORMAT:\\
\textless caption\textgreater 
\textless EXTRACTED SUBFIGURE CAPTION OR VERBATIM FULL CAPTION\textgreater
\textless /caption\textgreater
\end{tcolorbox}
\caption{An example prompt used for subcaption extraction with the Qwen2.5-VL-32B-Instruct model.}
\label{fig:subcap}
\end{figure*}

\begin{figure*}[t]
% \begin{tcolorbox}[
%     enhanced,
%     breakable,
%     colback=black!80,
%     coltext=white,
%     left=2mm, right=2mm, top=1mm, bottom=1mm,
%     boxrule=0pt,
%     sharp corners
% ]
% \textbf{\large Prompt: Figure-context summary generation}
% \end{tcolorbox}

% \begin{tcolorbox}[
%     enhanced,
%     breakable,
%     colback=gray!10,
%     colframe=gray!50,
%     left=3mm, right=3mm, top=2mm, bottom=2mm,
%     boxrule=0.3pt,
%     sharp corners
% ]
\begin{tcolorbox}[
    enhanced,
    % breakable,
    colback=gray!10,
    colframe=gray!50,
    colbacktitle=black!80,
    coltitle=white,
    title=\large \textbf{Prompt: Figure-context summary generation},
    left=3mm, right=3mm, top=1.5mm, bottom=1.5mm,
    boxrule=0.3pt,
    sharp corners,
]
\ttfamily
INSTRUCTIONS:\\
You will be provided with:\\
1. A subcaption that describes a subfigure from a compound figure.\\
2. The full caption of the compound figure.\\
3. A context passage related to the compound figure.\\
Definition of compound figure: A compound figure is a figure that contains multiple subfigures of the same topic (e.g., panels A, B, C, etc.).\\[1mm]
Your task is to summarize only the portions of the context passage that are most relevant to the given subcaption. The full caption is provided for additional information.\\
The summary should:\\
-- Use both the subcaption and the full caption to determine context.\\
-- Be concise and focused on the subcaption's content.\\
-- Exclude unrelated information from the context passage.\\
-- Preserve key biomedical terminology exactly as it appears.\\
-- Output the summary only, without any labels or additional text in the following format:\\
\textless summary\textgreater\\
\textless YOUR SUMMARY OF THE CONTEXT PASSAGE RELEVANT TO THE SUBCAPTION AND FULL CAPTION\textgreater\\
\textless /summary\textgreater
\end{tcolorbox}
\caption{An example prompt for figure-context summary generation using the Qwen2.5-14B-Instruct model.}
\label{fig:summary}
\end{figure*}

\section{Additional Subfigure Extraction Results}
\label{sec:add_sub_ext}
\begin{figure*}[t]
\centering
\includegraphics[width=0.99\textwidth]{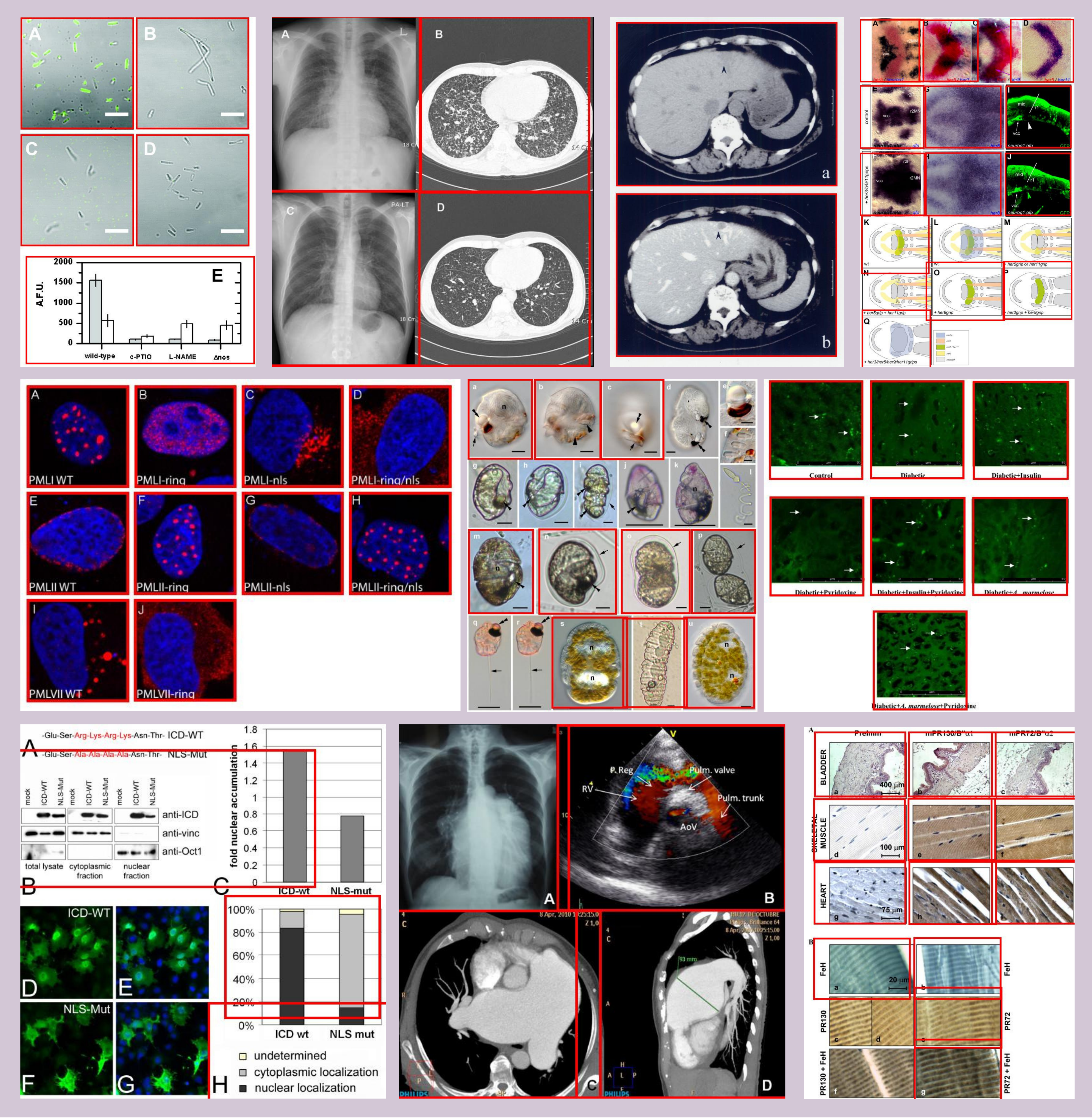}
\caption{Qualitative examples of subfigure detection using our DAB-DETR model. Shown are real-world biomedical compound figures from ImageCLEF 2016. Our model accurately localizes and separates distinct subfigures, even in heterogeneous, densely packed, or non-uniform layouts.}
\label{fig:example}
\end{figure*}

Figure~\ref{fig:example} showcases examples from the ImageCLEF 2016 dataset and from a subset of \datasetcomp, illustrating accurate detection of distinct subfigures across diverse panel layouts and content types.

\begin{figure*}[t]
\centering
\includegraphics[width=0.99\textwidth]{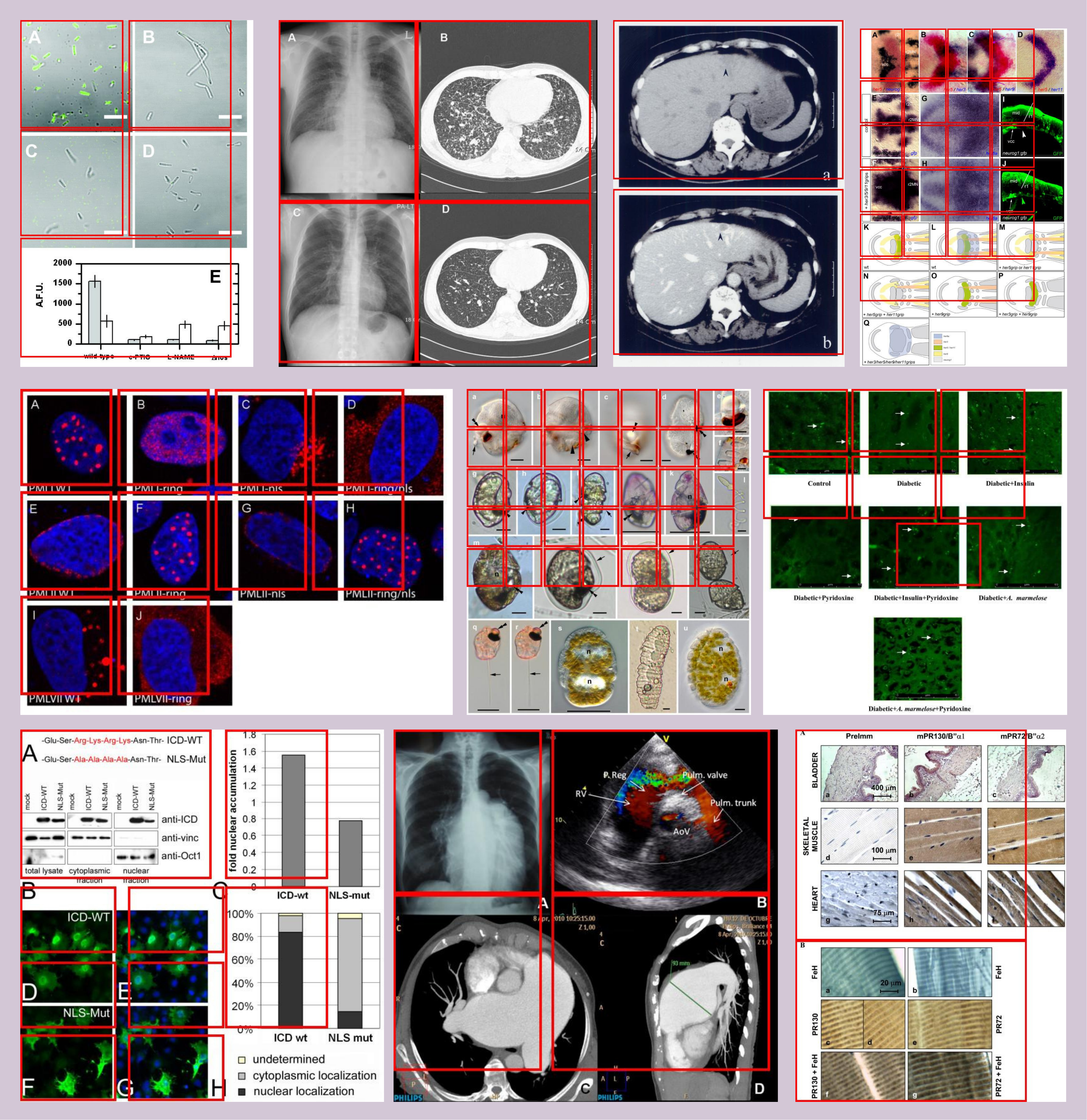}
\caption{Qualitative results of subfigure detection using Qwen2.5-VL-32B-Instruct on the same examples in Figure~\ref{fig:example} from ImageCLEF 2016. While Qwen can identify some coarse panel structure, it frequently under-segments, over-segments, or fails to localize several subfigures, especially in heterogeneous or irregular layouts, highlighting the advantages of our DAB-DETR model.}
\label{fig:qwenexample}
\end{figure*}

\section{Evaluation Results}
\subsection{Retrieval}
In addition to the results of recall at 200 which was included in the main body of the paper, we also provide the results for recall at 10 (Table \ref{ret-table-10}) and recall at 50 (Table \ref{ret-table-50}).

\subsection{Zeroshot Classification}
To assess the strength of the learned representations, we evaluate each model in a zero-shot classification setting. We use the model’s frozen image and text encoders without any task-specific training. Classification is done by comparing image features with the text prompts of each class and selecting the class with the highest similarity score. The results of the zero-shot experiments are shown in Table~\ref{tab:ZSC_F1}.
\begin{table*}[h!]
    \caption{Zero-shot classification F1-scores across diverse medical datasets for different models. Highest performance values are in bold, second-best are underlined.}
    \label{tab:ZSC_F1}
    \centering
    % \small
    \resizebox{\textwidth}{!}{
    \begin{tabular}{lcccccc}
    \toprule
      &\multicolumn{5}{c}{\textbf{Radiology}} \\
     \textbf{Model} & {\rotatebox{0}{\textbf{\fontsize{7}{7}\selectfont PneumoniaMNIST+}}} & {\rotatebox{0}{\textbf{\fontsize{7}{7}\selectfont BreastMNIST+}}} & {\rotatebox{0}{\textbf{\fontsize{7}{7}\selectfont OrganAMNIST+}}} & {\rotatebox{0}{\textbf{\fontsize{7}{7}\selectfont OrganCMNIST+}}} & {\rotatebox{0}{\textbf{\fontsize{7}{7}\selectfont OrganSMNIST+}}} & Average \\
    \midrule
    PMC-OA           & 50.94 & 52.36 & 19.70 & 14.79 & 16.99 & 30.95 \\
    \dataset         & 50.13 & 59.65 & 27.95 & 23.23 & 20.03 & \textbf{36.19} \\
    BioMedCLIP       & 60.13 & 33.76 & 19.40 & 14.12 & 16.00 & 28.62 \\
    BIOMEDICA        & 38.46 & 56.66 & 19.25 & 17.13 & 16.33 & 29.56 \\
    \midrule
    \datasetcomp     & 68.81 & 26.87 & 23.48 & 14.68 & 17.57 & 30.28 \\
    \datasetsub      & 38.46 & 61.18 & 28.43 & 24.28 & 20.38 & \underline{34.55} \\
    \end{tabular}
    }

    \resizebox{\textwidth}{!}{
    \begin{tabular}{lccccccc}
    \toprule
     &\multicolumn{6}{c}{\textbf{Visible Light 
 Photography}} \\
     \textbf{Model} & {\rotatebox{0}{\textbf{\fontsize{7}{7}\selectfont PAD-UFES-20}}} & {\rotatebox{0}{\textbf{\fontsize{7}{7}\selectfont Skin Cancer}}} & {\rotatebox{0}{\textbf{\fontsize{7}{7}\selectfont PathMNIST+}}} & {\rotatebox{0}{\textbf{\fontsize{7}{7}\selectfont DermaMNIST+}}} & {\rotatebox{0}{\textbf{\fontsize{7}{7}\selectfont OCTMNIST+}}} & {\rotatebox{0}{\textbf{\fontsize{7}{7}\selectfont RetinaMNIST+}}} & Average \\
    \midrule
    PMC-OA           & 17.18 & 13.30 & 56.03 & 14.29 & 50.74 & 27.22 & 29.79 \\
    \dataset           & 21.11 & 13.56 & 49.16 & 14.60 & 45.27 & 26.12 & 28.30\\
    BioMedCLIP        & 24.41 & 13.62 & 42.27 & 14.07 & 11.87 & 20.82 & 21.17 \\
    BIOMEDICA       & 40.57 & 17.20 & 49.10 & 21.89 & 10.00 & 18.53 & 26.21 \\
    \midrule
    \datasetcomp     & 33.04 & 16.56 & 52.17 & 17.52 & 46.91 & 22.81 & \underline{31.50} \\
    \datasetsub     & 27.00 & 17.64 & 66.02 & 14.16 & 37.71 & 30.09 & \textbf{32.10} \\
    \end{tabular}
    }

    \resizebox{\textwidth}{!}{
    \begin{tabular}{lccccccccc}
    \toprule
     &\multicolumn{8}{c}{\textbf{Microscopy}} \\
     \textbf{Model} & {\rotatebox{0}{\textbf{\fontsize{7}{7}\selectfont Sicap}}} & {\rotatebox{0}{\textbf{\fontsize{7}{7}\selectfont PCam}}} & {\rotatebox{0}{\textbf{\fontsize{7}{7}\selectfont NCT-CRC-HE}}} & {\rotatebox{0}{\textbf{\fontsize{7}{7}\selectfont LC-Lung}}} & {\rotatebox{0}{\textbf{\fontsize{7}{7}\selectfont LC-Colon}}} & {\rotatebox{0}{\textbf{\fontsize{7}{7}\selectfont BACH}}} &  {\rotatebox{0}{\textbf{\fontsize{7}{7}\selectfont BloodMNIST+}}} & {\rotatebox{0}{\textbf{\fontsize{7}{7}\selectfont TissueMNIST+}}} & Average \\
    \midrule
    PMC-OA           & 32.80 & 70.65 & 43.95 & 56.04 & 91.05 & 33.75 & 5.57 & 7.17 & 42.62\\
    \dataset          & 20.71 & 38.96 & 42.88 & 63.97 & 88.38 & 41.31 & 10.73 & 6.08 & 39.12 \\
    BIOMEDICA        & 31.80 & 62.17 & 48.98 & 70.93 & 84.43 & 39.83 & 4.37 & 4.31 & 43.35 \\
    BioMedCLIP       & 41.53 & 72.57 & 49.46 & 76.63 & 86.54 & 23.88 & 6.83 & 3.86 & 45.16 \\
    \midrule
    \datasetcomp     & 22.89 & 68.05 & 55.28 & 86.86 & 78.41 & 52.58 & 3.72 & 3.05 & \underline{46.35} \\
    \datasetsub      & 35.28 & 73.83 & 64.85 & 92.47 & 97.69 & 63.64 & 10.93 & 4.10 & \textbf{55.35} \\
    \bottomrule
    \end{tabular}
    }
    \label{zsc-tab}
\end{table*}

\subsection{Linear Probing}
To evaluate the quality of the learned image representations, we perform linear probing using a single-layer MLP on the downstream task datasets. Each model is trained for 40 epochs with a cosine annealing learning rate schedule, starting from an initial learning rate of 0.1. The results are presented in Table~\ref{tab:LP_F1}.

\begin{table*}[h!]
    \caption{Linear-probing F1-scores across diverse medical datasets for different models.}
    \label{tab:LP_F1}
    \centering
    % \small
    \resizebox{\textwidth}{!}{
    \begin{tabular}{lcccccc}
    \toprule
      &\multicolumn{5}{c}{\textbf{Radiology}} \\
     \textbf{Model} & {\rotatebox{0}{\textbf{\fontsize{7}{7}\selectfont PneumoniaMNIST+}}} & {\rotatebox{0}{\textbf{\fontsize{7}{7}\selectfont BreastMNIST+}}} & {\rotatebox{0}{\textbf{\fontsize{7}{7}\selectfont OrganAMNIST+}}} & {\rotatebox{0}{\textbf{\fontsize{7}{7}\selectfont OrganCMNIST+}}} & {\rotatebox{0}{\textbf{\fontsize{7}{7}\selectfont OrganSMNIST+}}} & Average \\
    \midrule
    BioMedCLIP       & 92.96 & 75.63 & 85.71 & 79.29 & 64.88 & 79.69 \\
    BIOMEDICA        & 86.15 & 77.16 & 89.72 & 82.66 & 70.93 & 81.12 \\
    \midrule
    \datasetcomp     & 79.74 & 77.84 & 89.56 & 85.00 & 69.07 & 80.24 \\
    \datasetsub      & 79.74 & 77.84 & 89.51 & 85.00 & 69.07 & 80.23 \\
    \end{tabular}
    }

    \resizebox{\textwidth}{!}{
    \begin{tabular}{lccccccc}
    \toprule
     &\multicolumn{6}{c}{\textbf{Visible Light 
 Photography}} \\
     \textbf{Model} & {\rotatebox{0}{\textbf{\fontsize{7}{7}\selectfont PAD-UFES-20}}} & {\rotatebox{0}{\textbf{\fontsize{7}{7}\selectfont Skin Cancer}}} & {\rotatebox{0}{\textbf{\fontsize{7}{7}\selectfont PathMNIST+}}} & {\rotatebox{0}{\textbf{\fontsize{7}{7}\selectfont DermaMNIST+}}} & {\rotatebox{0}{\textbf{\fontsize{7}{7}\selectfont OCTMNIST+}}} & {\rotatebox{0}{\textbf{\fontsize{7}{7}\selectfont RetinaMNIST+}}} & Average \\
    \midrule
    BioMedCLIP        & 62.31 & 56.43 & 90.27 & 59.62 & 71.70 & 42.95 & 63.88 \\
    BIOMEDICA         & 82.59 & 68.09 & 88.32 & 74.02 & 80.17 & 52.11 & 74.21 \\
    \midrule
    \datasetcomp     & 75.62 & 61.61 & 91.28 & 61.47 & 80.17 & 46.10 & 69.37 \\
    \datasetsub      & 75.62 & 62.92 & 91.35 & 61.47 & 78.73 & 46.59 & 69.44 \\
    \end{tabular}
    }

    \resizebox{\textwidth}{!}{
    \begin{tabular}{lccccccccc}
    \toprule
     &\multicolumn{8}{c}{\textbf{Microscopy}} \\
     \textbf{Model} & {\rotatebox{0}{\textbf{\fontsize{7}{7}\selectfont Sicap}}} & {\rotatebox{0}{\textbf{\fontsize{7}{7}\selectfont PCam}}} & {\rotatebox{0}{\textbf{\fontsize{7}{7}\selectfont NCT-CRC-HE}}} & {\rotatebox{0}{\textbf{\fontsize{7}{7}\selectfont LC-Lung}}} & {\rotatebox{0}{\textbf{\fontsize{7}{7}\selectfont LC-Colon}}} & {\rotatebox{0}{\textbf{\fontsize{7}{7}\selectfont BACH}}} &  {\rotatebox{0}{\textbf{\fontsize{7}{7}\selectfont BloodMNIST+}}} & {\rotatebox{0}{\textbf{\fontsize{7}{7}\selectfont TissueMNIST+}}} & Average \\
    \midrule
    BioMedCLIP       & 63.84 & 83.00 & 72.56 & 96.83 & 99.75 & 73.01 & 95.43 & 43.71 & 78.51 \\
    BIOMEDICA        & 65.15 & 86.41 & 83.57 & 99.26 & 99.95 & 75.65 & 96.92 & 50.69 & 82.2 \\
    \midrule
    \datasetcomp     & 60.00 & 84.22 & 64.64 & 98.85 & 99.80 & 62.39 & 95.87 & 49.89 & 76.95 \\
    \datasetsub      & 59.85 & 84.16 & 64.64 & 98.85 & 99.80 & 65.52 & 95.87 & 49.96 & 77.33 \\
    \bottomrule
    \end{tabular}
    }
\end{table*}
% \section{Rationale}
% \label{sec:rationale}
% % 
% Having the supplementary compiled together with the main paper means that:
% % 
% \begin{itemize}
% \item The supplementary can back-reference sections of the main paper, for example, we can refer to \cref{sec:intro};
% \item The main paper can forward reference sub-sections within the supplementary explicitly (e.g. referring to a particular experiment); 
% \item When submitted to arXiv, the supplementary will already included at the end of the paper.
% \end{itemize}
% 
% To split the supplementary pages from the main paper, you can use \href{https://support.apple.com/en-ca/guide/preview/prvw11793/mac#:~:text=Delete%20a%20page%20from%20a,or%20choose%20Edit%20%3E%20Delete).}{Preview (on macOS)}, \href{https://www.adobe.com/acrobat/how-to/delete-pages-from-pdf.html#:~:text=Choose%20%E2%80%9CTools%E2%80%9D%20%3E%20%E2%80%9COrganize,or%20pages%20from%20the%20file.}{Adobe Acrobat} (on all OSs), as well as \href{https://superuser.com/questions/517986/is-it-possible-to-delete-some-pages-of-a-pdf-document}{command line tools}.

\end{document}